%% file: main_tech_report.tex
\documentclass[10pt, logo, copyright]{nvidiatechreport}

\usepackage{mdframed}
\usepackage[utf8]{inputenc} 
\usepackage[T1]{fontenc}    

\usepackage{amsfonts}       
\usepackage{nicefrac}       
\usepackage{microtype}      
\usepackage[dvipsnames]{xcolor}         
\usepackage{multirow}
\usepackage{multicol}
\usepackage{graphicx}
\usepackage[numbers]{natbib}
\usepackage{tabto}
\usepackage{xspace}
\usepackage{amsmath}
\usepackage{adjustbox}
\usepackage{enumitem}
\usepackage{wrapfig}
\usepackage{dblfloatfix}

\usepackage{footmisc}
\usepackage{listings}

\usepackage{float}

\usepackage{hyperref}
\usepackage{array}
\usepackage{ragged2e}
\usepackage{subcaption}
\usepackage{caption}

\usepackage{natbib}
\input{math_commands.tex}

\usepackage{amsmath}
\usepackage{amssymb}
\usepackage{mathtools}
\usepackage{amsthm}
\usepackage{enumitem}
\setlist[itemize]{itemsep=0.5em}
\usepackage{multirow}
\usepackage{booktabs}

\usepackage[capitalize,noabbrev]{cleveref}

\usepackage{algorithm}
\usepackage{algorithmicx}
\usepackage{algpseudocode}
\usepackage{xspace}
\usepackage{csquotes}
\usepackage{listings}
\usepackage{xcolor}

\usepackage{multirow}
\usepackage{multicol}
\usepackage{enumitem}
\usepackage[most,skins,theorems]{tcolorbox}
\tcbset{
  aibox/.style={
    width=\linewidth,
    top=8pt,
    bottom=4pt,
    colback=blue!6!white,
    colframe=black,
    colbacktitle=black,
    enhanced,
    center,
    attach boxed title to top left={yshift=-0.1in,xshift=0.15in},
    boxed title style={boxrule=0pt,colframe=white,},
  }
}

\newtcolorbox{AIbox}[2][]{aibox,title=#2,#1}

\definecolor{lightgray}{gray}{0.95} 
\definecolor{darkblue}{rgb}{0,0,0.6} 
\definecolor{nvgreen}{cmyk}{50, 0, 100, 0}

\title{Scaling Up RL: Unlocking Diverse Reasoning in LLMs via Prolonged Training}

\author{NVIDIA}

\correspondingauthor{}

\newcommand\shizhe[1]{{\color{orange}[#1]$_{Shizhe}$}}

\begin{abstract}
Recent advancements in reasoning-focused language models such as OpenAI’s O1 and DeepSeek-R1 have shown that scaling test-time computation-through
chain-of-thought reasoning and iterative exploration—can yield substantial improvements on complex tasks like mathematics and code generation. These breakthroughs have been driven by large-scale reinforcement learning (RL), particularly when combined with verifiable reward signals that provide objective and grounded supervision. In this report, we investigate the effects of prolonged reinforcement learning on a small language model across a diverse set of reasoning domains. Our work identifies several key ingredients for effective training, including the use of verifiable reward tasks, enhancements to Group Relative Policy Optimization (GRPO), and practical techniques to improve training stability and generalization. We introduce controlled KL regularization, clipping ratio, and periodic reference policy resets as critical components for unlocking long-term performance gains. Our model achieves significant improvements over strong baselines, including +14.7\% on math, +13.9\% on coding, and +54.8\% on logic puzzle tasks. To facilitate continued research, we release our model publicly.
\vspace{2mm}
\newline
\textbf{Model on Hugging Face:} \href{https://huggingface.co/nvidia/Nemotron-Research-Reasoning-Qwen-1.5B}{Nemotron-Research-Reasoning-Qwen-1.5B}
\end{abstract}

\begin{document}
\maketitle
\section{Introduction}

The emergence of advanced reasoning models such as OpenAI's O1~\citep{jaech2024openai} and DeepSeek-R1~\citep{guo2025deepseek} represents a paradigm shift in Large Language Models (LLMs). This shift is characterized by scaling test-time computation-enabling models to perform extended deliberation through long-form Chain-of-Thought (CoT) reasoning. By allocating computational resources on 
using different reasoning strategies such as
exploration, verification, and backtracking during the generation process, these models achieve remarkable improvements on complex tasks such as solving mathematical problems and code generation.


This paradigm shift has been primarily enabled by large-scale Reinforcement Learning (RL), which unlocks sophisticated reasoning capabilities beyond what supervised 
finetuning
alone can achieve. The key innovation driving these advances is the use of tasks with verifiable 
rewards
where correctness can be programmatically verified.
This approach provides reliable training signals without the vulnerabilities inherent to learned reward models, which often suffer from reward hacking (such as generating superficially polite or unnecessarily verbose responses). Instead, verifiable rewards deliver grounded feedback that directly corresponds to correct reasoning and successful task completion~\citep{guo2025deepseek, deepscaler2025}.

In this technical report, we investigate how prolonged reinforcement learning affects model performance, stability, and generalization across diverse reasoning tasks. Despite recent progress in this area, challenges remain in developing stable, scalable and sound training recipes for reasoning language models. These challenges include entropy collapse during training, difficulty in maintaining exploration-exploitation balance, and performance plateaus during extended optimization. Through systematic experimentation, we identify critical components that collectively form an effective framework for large-scale reinforcement learning of language models:

\textbf{Diverse Training Data.} We scale training across a wide variety of tasks with verifiable reward signals, spanning traditional reasoning domains such as mathematics and code generation, to more complex areas including STEM problem solving, logical puzzles, and instruction following. This data diversity helps evaluate the generality of reinforcement learning algorithms and exposes the model to a broader distribution of reasoning strategies. Compared to \textit{DeepSeek-R1-Distill-Qwen-1.5B}, we achieve average pass@1 improvements of 14.7\% on math benchmarks, 13.9\% on coding, 54.8\% on logic puzzles, 25.1\% on STEM reasoning, and 18.1\% on instruction-following.

\textbf{Improvements over GRPO.} Through extensive ablation studies, we incorporate and validate several enhancements to Group Relative Policy Optimization (GRPO) proposed in DAPO~\cite{yu2025dapoopensourcellmreinforcement}, including decoupled clipping and dynamic sampling. These modifications lead to more efficient policy learning, with observed improvements as much as +5\% on AIME2024 validation scores in our small-scale ablation study. In particular, dynamic sampling increases effective sample efficiency, and relaxed clipping coefficients over the importance sampling ratio contributes to the stabilization of entropy.

\textbf{Training Stability.} We found the training procedure to be sensitive to several critical hyperparameters, especially those related to sampling temperature, clipping thresholds, and KL divergence. While recent work suggests removing KL regularization, we find it beneficial in balancing exploration and exploitation, particularly when training from strong pretrained models. Applying a small but non-zero KL penalty, combined with entropy-preserving strategies such as those from DAPO, stabilizes learning and prevents entropy collapse. These adjustments enable scalable and stable training over hundreds of thousands of steps.

\textbf{Prolonged Training.}
To support continued improvement beyond early convergence, we periodically reset the reference policy and optimizer states during training. Maintaining a static reference can restrict learning by penalizing beneficial divergence from the base model. Our results show that strategically resetting the reference policy, especially after observing KL spikes or validation performance declines, restores learning dynamics and enables further performance gains. For example, a mid-training reset recovered coding benchmark performance after a sharp decline, illustrating the effectiveness of this strategy in mitigating stagnation.

To support further research and development in this area, we are open-sourcing our model and hope that our contributions will be valuable to the broader research community.

\section{Diverse Training Data}
\label{sec:data}
We scale training across a wide spectrum of tasks that provide verifiable reward signals as show in~\Cref{tab:data_details}. 
These tasks span from traditional reasoning domains, such as mathematical problem solving and code generation, to more complex and open-ended domains, including STEM-related problem solving, logical puzzles, and instruction following. 
The inclusion of such a diverse task set serves two key purposes. First, it broadens the model’s exposure to a wide distribution of reasoning patterns, encouraging generalization beyond narrow, domain-specific behaviors. This is especially critical for developing models of adapting to new or unseen task formulations. Second, the task diversity enables a more rigorous evaluation of RL algorithms, as it tests their ability to learn robust decision-making strategies across fundamentally different environments and reward structures. By grounding training in tasks with clear correctness criteria, we ensure that observed improvements are attributable to genuine progress in reasoning ability, rather than spurious correlations or overfitting to specific task formats.

\begin{table}[t]
\centering
\begin{tabular}{|l|c|c|>{\RaggedRight\arraybackslash}p{4.5cm}|}
\hline
\textbf{Data Type} & \textbf{Reward Type} & \textbf{Quantity} & \textbf{Data Source} \\
\hline
Math        & Binary     & 40k   & \href{https://huggingface.co/datasets/agentica-org/DeepScaleR-Preview-Dataset}{DeepScaleR Dataset} \\
Code        & Continuous & 24k   & \href{https://huggingface.co/datasets/PRIME-RL/Eurus-2-RL-Data}{Eurus-2-RL Dataset} \\
STEM        & Binary     & 25k   & \href{https://huggingface.co/datasets/EricLu/SCP-116K}{SCP-116K Dataset} \\
Logical Puzzles & Continuous & 37k & \href{https://github.com/open-thought/reasoning-gym}{Reasoning Gym} \\
Instruction Following & Continuous & 10k & \href{https://huggingface.co/datasets/nvidia/Llama-Nemotron-Post-Training-Dataset/viewer/RL}{Llama-Nemotron} \\
\hline
\end{tabular}
\caption{Overview of training data used in our experiments, categorized by domain, reward type (binary or continuous), dataset size, and source. The datasets span a range of reasoning, coding, STEM, and instruction-following tasks, sourced from public repositories.}
\label{tab:data_details}
\end{table}

\subsection{Math}
We use high-quality, community-curated datasets made available through DeepScaleR~\citep{deepscaler2025}.
The training set consists of 40K math problems sourced from a diverse range of national and international math competitions. 
We adopt DeepScaleR's original verifier and further augment it with an improved \texttt{math-verify}\footnote{\url{https://github.com/huggingface/Math-Verify}}. 
We obtain the LLM's answers by prompting the model with \texttt{Let's think step by step and output the final answer within \textbackslash boxed\{\}.} 
We use a binary reward signal, assigning a score of 1 if the LLM's response passes either the original or the enhanced \texttt{math-verify}, and 0 otherwise (for incorrect or improperly formatted answers).

\subsection{Code}
We utilize publicly available reinforcement learning datasets comprising 24K coding problems~\cite{cui2025process}, sourced from various programming competitions. To support continuous reward feedback, we improve code execution environment to run all test cases rather than terminating on the first error and assign rewards based on the fraction of test cases passed. Submissions that fail to compile, contain syntax errors, or exceed a 5 second total timeout are assigned a reward of zero. We also include instructions for the LLM to enclose its final code response with triple backticks. 

\subsection{STEM}

We use SCP-116K~\citep{lu2025scp116khighqualityproblemsolutiondataset}, a large-scale dataset containing 274K scientific problem-solution pairs spanning diverse fields such as physics, chemistry, biology, and mathematics.
Each problem is accompanied by a corresponding solution extracted from the original source text, along with model-generated responses and reasoning paths produced by DeepSeek-R1.
Given that SCP-116K was automatically extracted from heterogeneous and potentially noisy sources, we applied rigorous data filtering.
First, we removed problems lacking a retrievable ground-truth solution from the source text.
Then, we employed GPT-4o as a judge to assess whether the DeepSeek-R1 response aligned with the ground-truth answer.
Only problems with consistent answers were retained, reducing the dataset from the original entries to 25K.

\subsection{Logical Puzzles}

The logical puzzles are well-suited for reasoning model training due to their broad coverage of different reasoning skills, as well as their clear objectives and evaluation metrics. We utilize the Reasoning Gym project~\cite{stojanovski2025reasoninggymreasoningenvironments}, which offers approximately 100 tasks across various domains, including algebra, arithmetic, computation, cognition, geometry, graph theory, logic, and popular games. To facilitate model training and evaluation, we generate a large dataset consisting of 37K synthetic training samples and 9600 validation samples, spanning 96 tasks. Notably, some tasks have a unique solution, whereas others, such as the Rubik's Cube and Countdown, admit multiple correct solutions. We employ the verifier provided by the Reasoning Gym repository for both model evaluation and reinforcement learning training signals. We use recommended default prompts which instruct models to enclose answers between \texttt{<answer> </answer>} tags.

\subsection{Instruction Following}

To enhance our model's instruction-following capabilities, we leverage synthetic generated data~\cite{nvidia2024nemotron4340btechnicalreport} which data format is similar to IFEval~\cite{zhou2023instructionfollowingevaluationlargelanguage}. Specifically the dataset contains synthetic prompts that pair tasks with randomly chosen instructions. For instance, a prompt may ask the model to ``Write an essay about machine learning'', while the instruction specifies, ``Your response should have three paragraphs.'' We do not add further instructions on formatting and obtain the models response after thinking (\texttt{</think>} token). For evaluation, we adopt the strict mode of the IFEval verifier, which assesses only the original response and is considered more rigorous than the loose mode.

\subsection{Implementation Details}
To accommodate the complexity and diversity of our reward-generating tasks, we adopt a sandboxed reward server architecture. This design is motivated by several key factors. First, the diversity of data sources and task formats necessitates customized execution environments that can be tailored per task without interfering with the core training pipeline. Second, sandboxing allows us to isolate code execution from the training process, which is critical for security and fault-tolerance, especially when evaluating code generation or interacting with external systems. This ensures that any potential side effects or crashes during reward computation do not compromise the training run. Finally, we leverage multiprocessing and distributed reward servers across training clusters to scale reward evaluation efficiently. This parallelization is essential for maintaining high throughput during reinforcement learning, especially when reward signals to compute or involve latency-bound operations such as code execution. We also launch reward calculations as soon as responses are completed, overlapping reward calculation with GPU tasks which do not depend on reward availability, such as $\pi_\theta(\tau), \pi_{old}(\tau)$ calculations.

\input{main_method}
\section{Experiment Results}

To ensure stable and effective reinforcement learning across diverse reasoning tasks, we adopt a staged training strategy involving multiple sequential runs, each designed to address specific challenges observed in earlier phases. This approach allows us to iteratively refine model behavior, incorporate additional data sources, adjust hyperparameters, and reset training dynamics when necessary. Throughout, we closely monitor key training signals, including KL divergence, entropy, and response length, and scores on a held-out validation set, using them as indicators to guide interventions such as reward shaping or context window changes. Our complete training recipe reflects this progressive, intervention-driven process, which we detail below.

\subsection{Training Setup}
\label{sec:hyperparams}
We primarily levearge open-source framework verl~\cite{Sheng_2025} for reinforcement learning training. 
We adopt proposed enhancements from DAPO~\cite{yu2025dapoopensourcellmreinforcement}, decoupling clipping hyperparameters with $\epsilon_{low}=0.2, \epsilon_{high}=0.4$, and dynamic sampling for filtering prompts with accuracy equal to 1 and 0. 
We apply KL divergence penalty with a small penaly coefficient $\beta=0.0001$. 
For rollout, we sample $n=16$ responses for each prompt with a context window limit of 8096 and use a high sampling temperature of 1.2. 
We set batch size to 256 and mini-batch size to 64 (equating to 4 gradient updates per rollout step). For training we use the AdamW~\cite{loshchilov2019decoupledweightdecayregularization} optimizer with a constant learning rate of $2\times 10^{-6}$.
We use \textit{DeepSeek-R1-Distill-Qwen-1.5B} as the initial base model policy.
We conduct training on 4 8 x NVIDIA-H100-80GB nodes, and the whole training runs for approximately 16k GPUs hours.

\subsection{Training Recipe}

\begin{figure}[ht] 
    \centering
    \begin{subfigure}[b]{0.48\textwidth}
        \includegraphics[width=\textwidth]{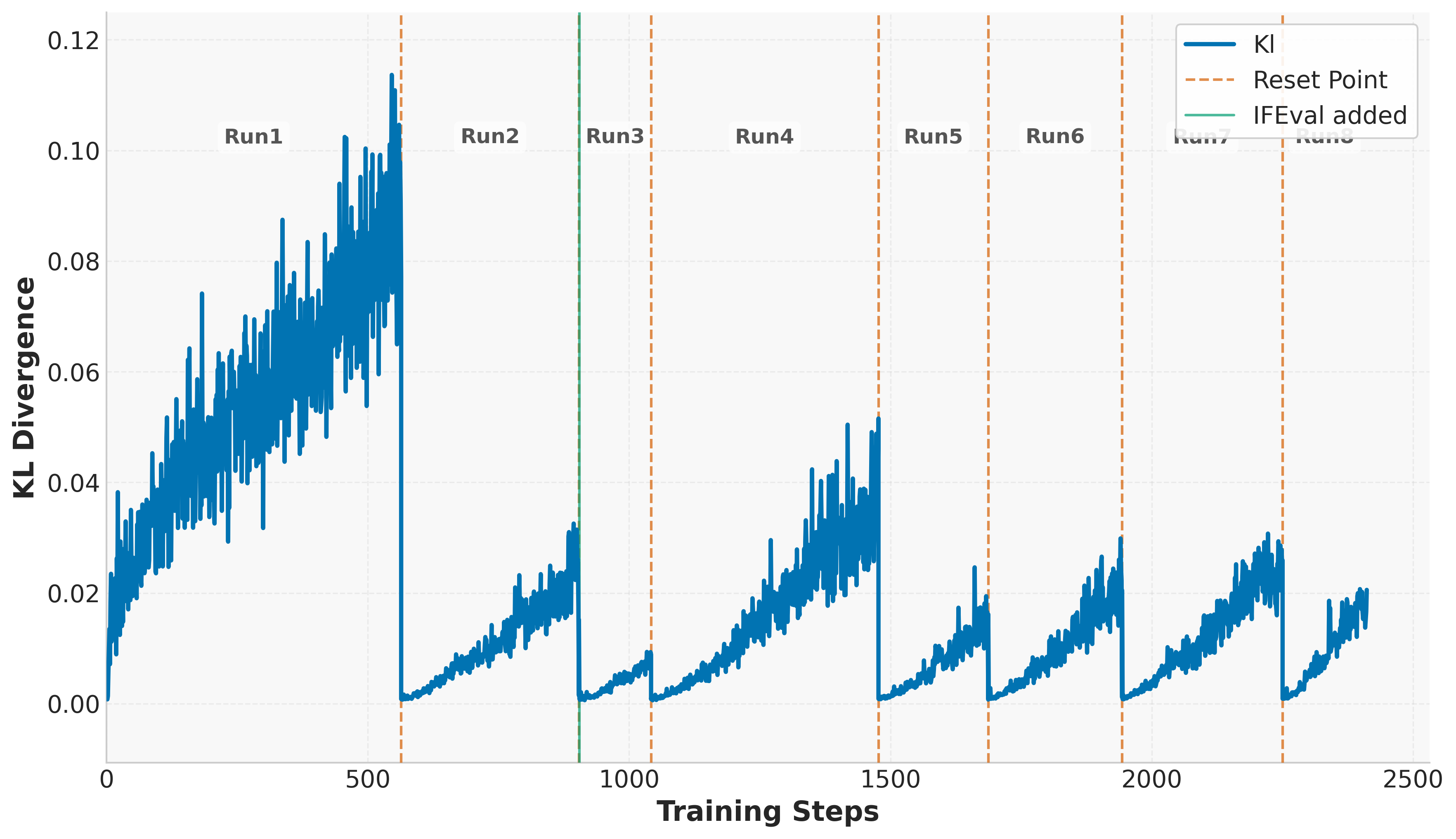}
        \caption{KL divergence.}
    \end{subfigure}
    \hfill
    \begin{subfigure}[b]{0.48\textwidth}
        \includegraphics[width=\textwidth]{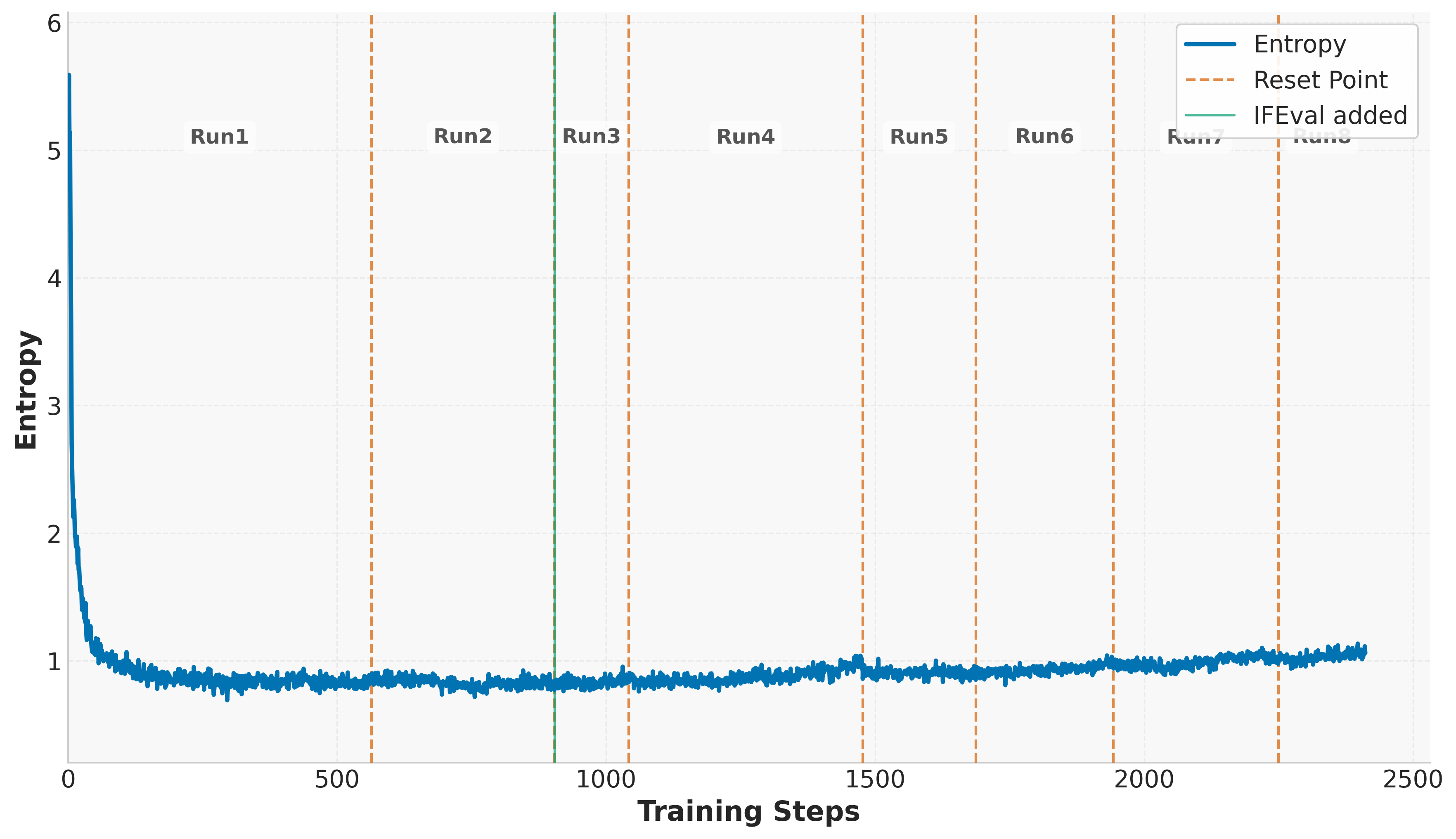}
        \caption{Entropy.}
    \end{subfigure}

    \vspace{0.5cm}

    \begin{subfigure}[b]{0.48\textwidth}
        \includegraphics[width=\textwidth]{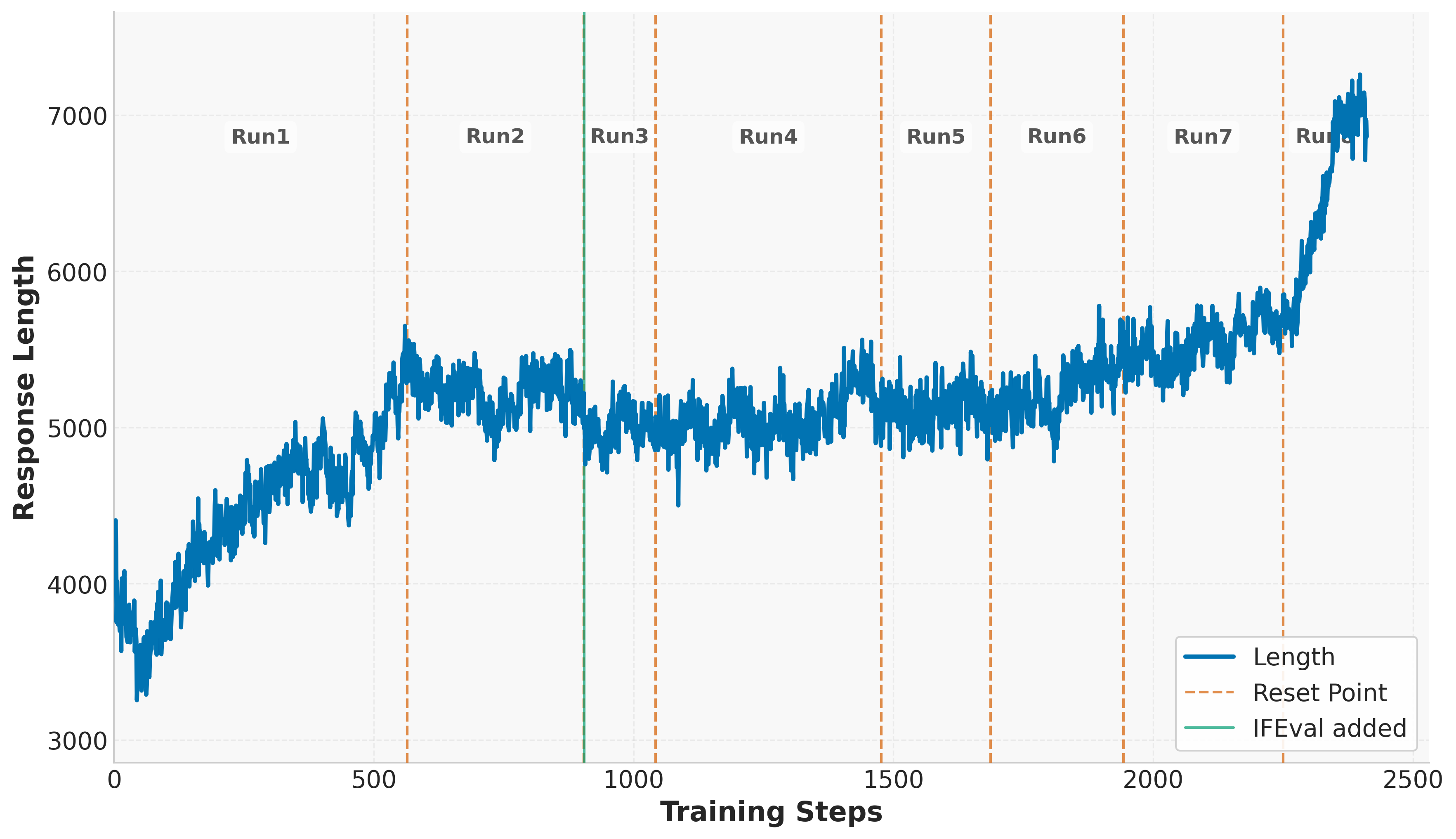}
        \caption{Response length.}
    \end{subfigure}
    \hfill
    \begin{subfigure}[b]{0.48\textwidth}
        \includegraphics[width=\textwidth]{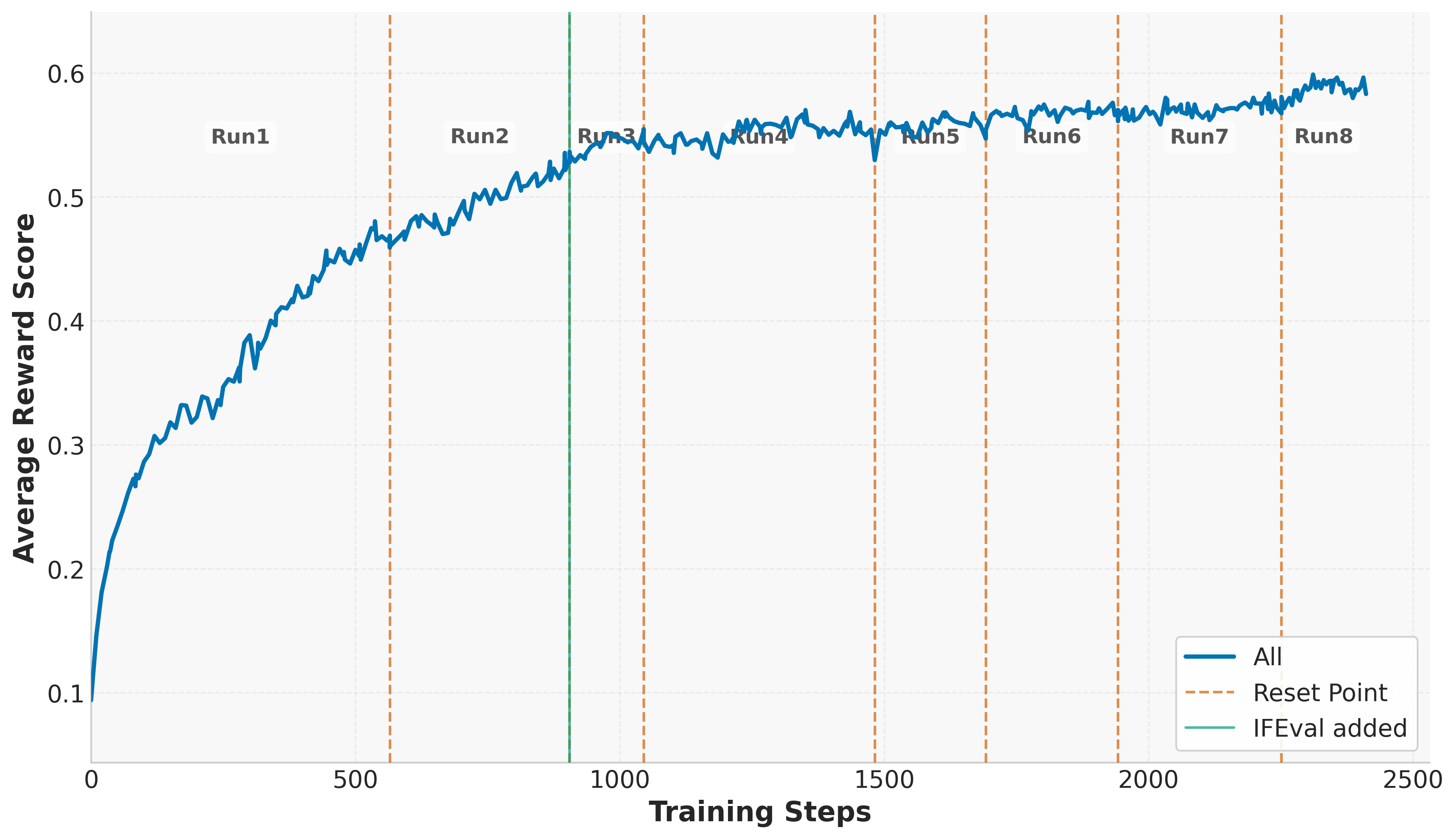}
        \caption{Averaged validation scores.}
    \end{subfigure}

    \caption{Training dynamics across multiple runs. This figure presents key metrics monitored during training, including KL divergence, entropy, response length, and validation performance across different stages. Each run reflects specific adjustments to training conditions, such as changes in context window size, reward shaping, and rollout counts. The dynamics highlight the impact of these adjustments on model stability, response quality, and performance improvements on validation scores.}
    \label{fig:training_dynamic}
\end{figure}

We construct a validation data blend to closely monitor training progress across steps. This validation set includes subsets from our evaluation benchmark, specifically AIME2024, Codeforces, GPQA-diamond, IFEval, and the logic puzzle \textit{graph\_color} from Reasoning Gym. We evaluate model performance using similar sampling parameters as in evaluation settings other than the same context window used in training. Occasionally, we perform a hard reset of the reference model, as described in~\Cref{sec:kl_reset}, particularly when validation metrics significantly degrade or when improvements plateau. Unless otherwise specified, we use the hyperparameters detailed in~\Cref{sec:hyperparams}.

Interestingly, the hard reset (resetting reference policy and optimizer states) not only restores training stability but also provides an opportunity to adjust training hyperparameters and introduce enhancements such as additional training data and reward shaping. \Cref{fig:training_dynamic} presents key statistics on training dynamics across different stages. The final training recipe comprises several sequential stages, described below:

\begin{itemize}
    \item \textbf{Run 1:} We begin training on four tasks from~\Cref{sec:data}, excluding instruction-following data. In this phase, the model initially adapts to a reduced context window of 8k, leading to shorter responses. Subsequently, response length increases along with improved validation scores. Toward the end of this stage, we observe instability and degradation in validation performance.
    
    \item \textbf{Run 2:} We perform a hard reset of the reference policy and resume training with the same setup as Run 1. Unlike DeepScaleR~\cite{deepscaler2025}, which proposes increasing the context window, we maintain a fixed 8k context. Notably, validation scores continue to improve in this stage, even as response length remains relatively stable.

\item \textbf{Run 3:} We incorporate instruction-following data into the training mix and continue training. This stage proceeds until we observe a sudden increase in response length, primarily due to the model repeating answers and failing to terminate with an \textit{eos} token.

\item \textbf{Runs 4 and 5:} We introduce reward shaping by penalizing responses that do not terminate correctly. This encourages proper generation behavior, resulting in a modest reduction in response length. Toward the end of this stage, gains on validation benchmarks begin to plateau.

\item \textbf{Runs 6 and 7:} We increase the rollout count from 16 to 32, performing two hard resets in the process. Interestingly, response length begins to rise again alongside improvements in validation metrics.

\item \textbf{Run 8:} We extend the context window to 16k tokens and reduce rollout count to 16. Despite the model being trained on an 8k context window for most of the time, it quickly adapts to the extened context window. We observe marginal improvements in hard math tasks like AIME, with more substantial gains coming from other domains.
\end{itemize}

\begin{figure}[t] 
    \centering
    \includegraphics[width=0.95\textwidth]{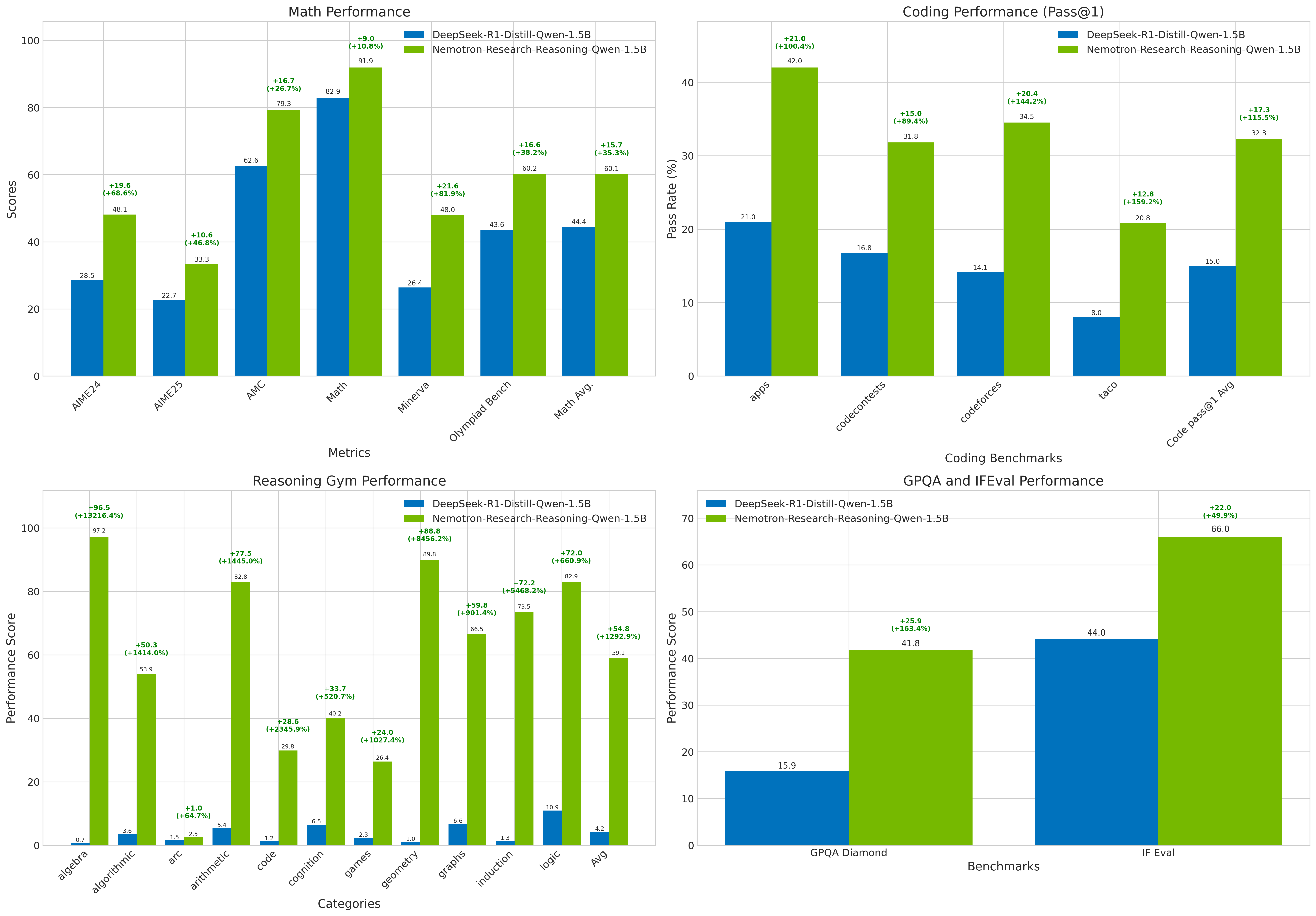}
    \caption{Performance comparison between \textit{DeepSeek-R1-Distill-Qwen-1.5B} and our model \textit{Nemotron-Research-Reasoning-Qwen-1.5B} across multiple benchmarks in diverse domains, including math, coding, Reasoning Gym puzzles, GPQA Diamond, and IFEval.}
    \label{fig:deepseek_compare}
\end{figure}

\subsection{Evaluation Benchmarks}

We evaluate models on the breadth of various tasks across math, coding, reasoning, and instruction following.
For math, we follow DeepScaleR~\cite{deepscaler2025} and SimpleRL~\cite{zeng2025simplerl}, and evaluate on AIME2024~\cite{AIME2024}, AIME2025~\cite{AIME2025}, AMC~\cite{AMC} (composed of AMC2022 and AMC2023), MATH~\cite{hendrycks2021measuringmathematicalproblemsolving}, Minerva Math~\cite{lewkowycz2022solvingquantitativereasoningproblems}, and Olympiad Bench~\cite{he2024olympiadbenchchallengingbenchmarkpromoting}. 
For coding, we use the validation set from PRIME~\cite{cui2025process} consisted of APPS~\cite{hendrycks2021measuringcodingchallengecompetence}, Codecontests~\cite{Li_2022}, Codeforces\footnote{\url{https://huggingface.co/datasets/MatrixStudio/Codeforces-Python-Submissions}}, and TACO~\cite{li2023tacotopicsalgorithmiccode}. 
We also include benchmarks HumanevalPlus~\cite{evalplus} and LiveCodeBench~\cite{jain2024livecodebenchholisticcontaminationfree}.
For logic puzzles, we reserved 100 samples from each reasoning gym~\cite{stojanovski2025reasoninggymreasoningenvironments} tasks as test datasets for evaluation.
In addition, we use a curated subset\footnote{\url{https://huggingface.co/datasets/spawn99/GPQA-diamond-ClaudeR1}} from GPQA Diamond~\cite{rein2023gpqagraduatelevelgoogleproofqa} and IFEval~\cite{bae2025onlinedifficultyfilteringreasoning} to evaluate the capability of our models in STEM reasoning and instruction following~\cite{zhou2023instructionfollowingevaluationlargelanguage}. 

\subsection{Evaluation Results}
We use vllm (version 0.7.2)~\cite{kwon2023efficient} for inference backend, with sampling temperature of 0.6, nucleus sampling~\cite{holtzman2020curiouscaseneuraltext} with $top\_p=0.95$ and a context windo of 32k. For math, coding and STEM reasoning tasks we obtain estimates of $pass@1$ from 16 samples for each benchmark prompts from strictly binary rewards. For other tasks such as logical puzzles and instruction following, we calculate the average continuous reward score from our rule-based verifiers with 16 samples. We evaluate and report benchmark results for open-source models using our own evaluation settings. We acknowledge that discrepancies from originally reported numbers may arise due to differences in evaluation settings and inherent variability from inference.

\Cref{fig:deepseek_compare} provides a detailed comparison between \textit{DeepSeek-R1-Distill-Qwen-1.5B} and our model, \textit{Nemotron-Research-Reasoning-Qwen-1.5B}, across multiple domains: math, coding, reasoning, and instruction following. In the math domain, our model consistently outperforms across benchmarks, showing an average improvement of 15.7\%, which reflects enhanced symbolic and numerical reasoning capabilities. In coding, \textit{Nemotron-Research-Reasoning-Qwen-1.5B} surpasses \textit{DeepSeek-R1-Distill-Qwen-1.5B} in competitive programming tasks as measured by $pass@1$ accuracy, indicating superior code synthesis ability. Our model also demonstrates substantial gains in STEM reasoning and instruction following, with improvements of 25.9\% on GPQA Diamond and 22.0\% on IFEval. For logic puzzles in the Reasoning Gym suite, we adopt the categorization of 96 tasks as defined by the official GitHub repository. Notably, \textit{DeepSeek-R1-Distill-Qwen-1.5B} underperforms even on relatively simple mathematical tasks such as algebra and arithmetic. Closer inspection reveals that the model consistently formats its answers using \texttt{\textbackslash boxed\{\}} rather than adhering to the instruction to use \texttt{<answer> </answer>} tags. Despite poor initial formatting behavior, the model is able to achieve high accuracy on these easier tasks post training, suggesting that formatting is relatively easy to learn. Our models still exhibit room for improvement on more challenging categories, including tasks from arc, code, cognition, and games. In these cases, the model often fails to make meaningful progress. Further analysis indicates that these failures stem from either a lack of core reasoning skills necessary to solve specific subtasks or insufficient background knowledge related to the problem domains. Addressing these limitations may require additional finetuning data to better support model from a cold start, which we leave these enhancements to future work.

\begin{figure}[t] 
    \centering
    \begin{subfigure}[b]{0.48\textwidth}
        \includegraphics[width=\textwidth]{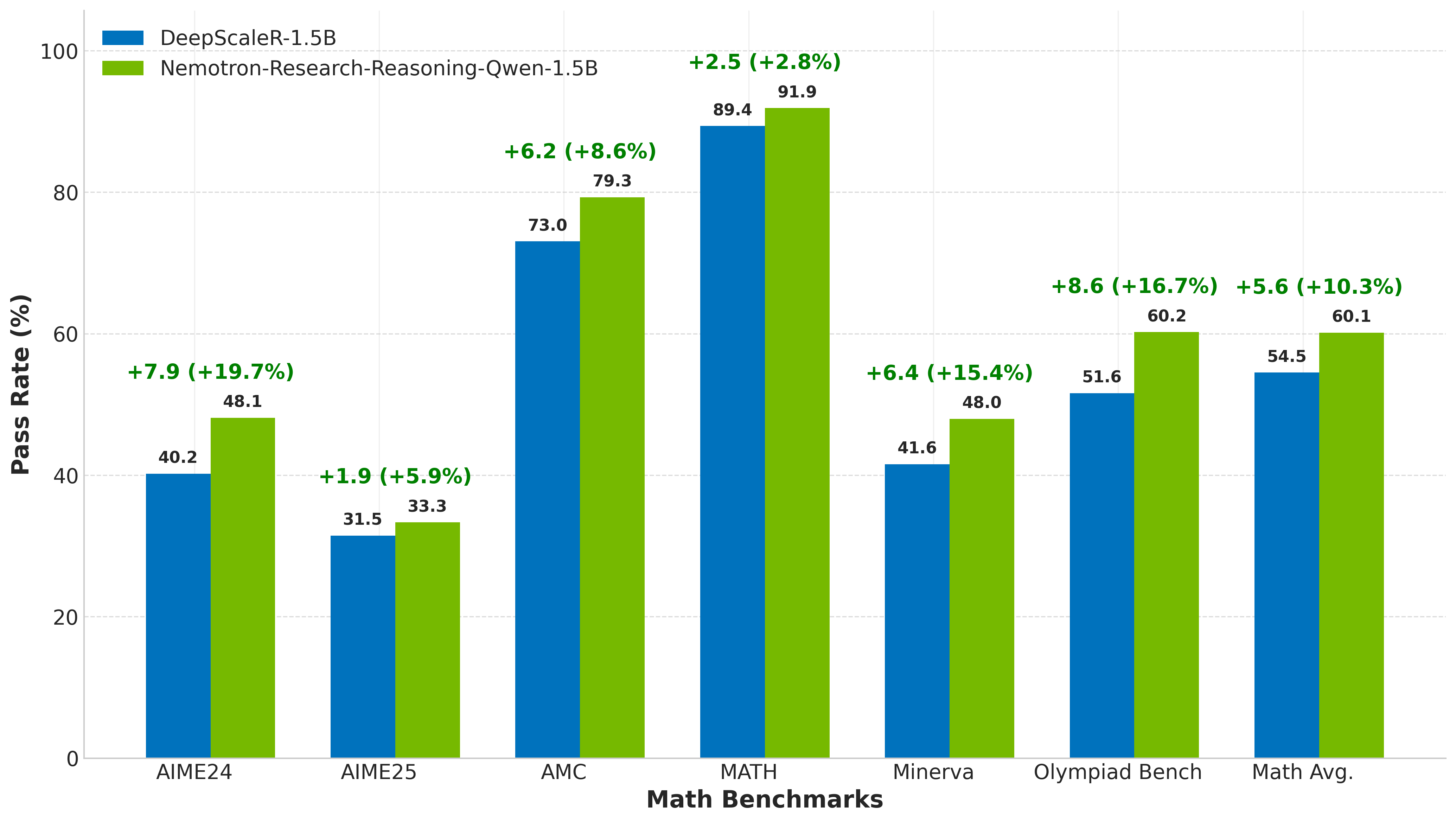}
        \caption{Comparision with DeepScaleR-1.5B~\cite{deepscaler2025}.}
    \end{subfigure}
    \hfill
    \begin{subfigure}[b]{0.48\textwidth}
        \includegraphics[width=\textwidth]{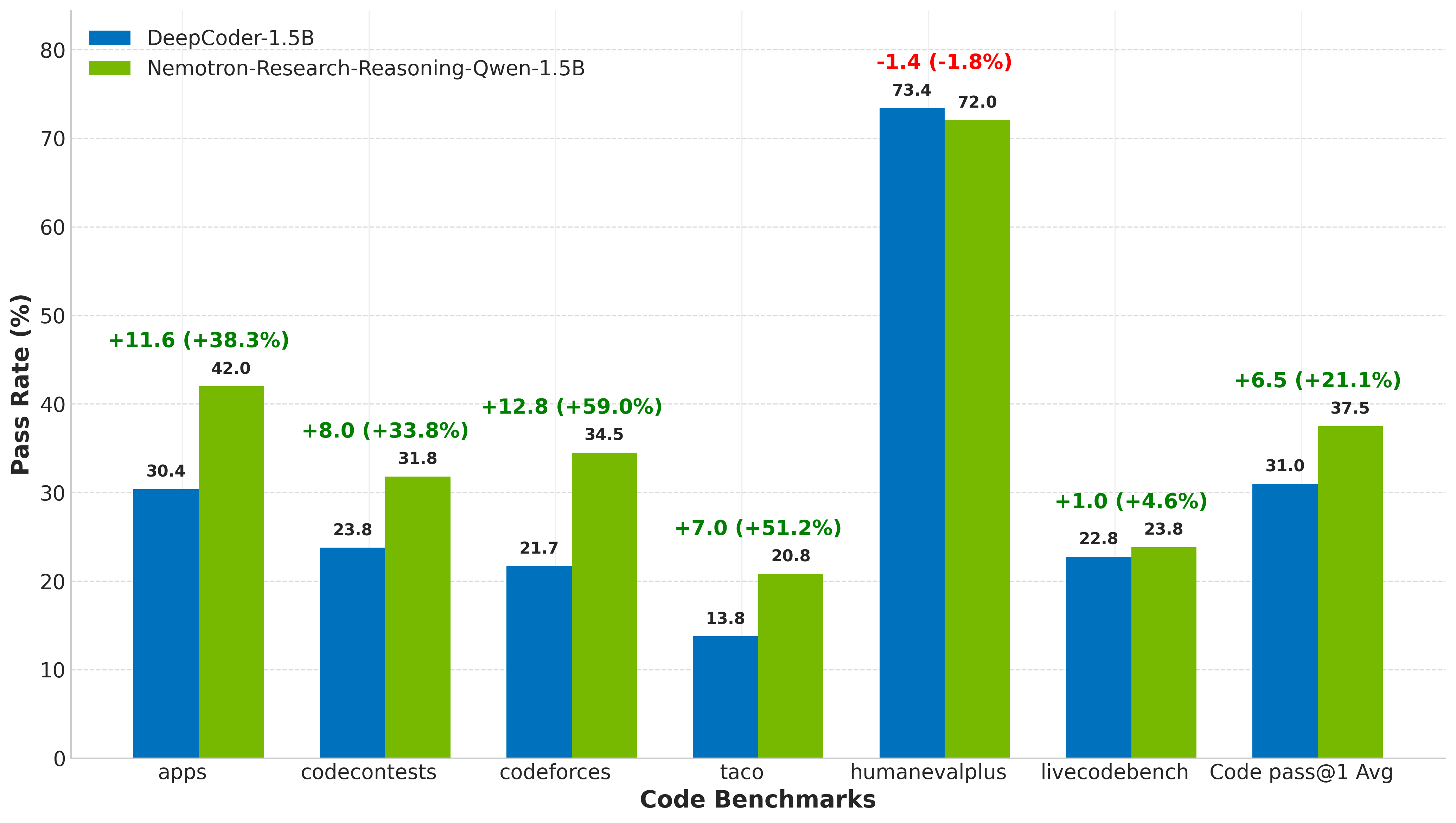}
        \caption{Comparision with DeepCoder-1.5B~\cite{deepcoder2025}.}
    \end{subfigure}

    \caption{Performance comparison of \textit{Nemotron-Research-Reasoning-Qwen-1.5B} with domain-specialized models \textit{DeepScaleR-1.5B} and \textit{DeepCoder-1.5B}. Trained on diverse-domain data, our model achieves competitive results in both mathematical reasoning and code generation tasks.}
    \label{fig:deepscalercoder_comparision}
\end{figure}

We further compare the performance of \textit{Nemotron-Research-Reasoning-Qwen-1.5B} against two domain-specialized baselines, \textit{DeepScaleR-1.5B} (optimized for mathematical reasoning) and \textit{DeepCoder-1.5B} (focused on competitive programming tasks) in~\Cref{fig:deepscalercoder_comparision}. Trained on a broad and diverse set of domains rather than being narrowly specialized, our model demonstrates strong generalization capabilities and achieves competitive performance across both math and code benchmarks.

\section{Ablation Studies}
To better understand the contribution of key components in our training pipeline, we conduct a series of ablation studies. Specifically, we examine the effects of rollout temperature sampling, decoupled clipping and dynamic sampling enhancements proposed in DAPO~\cite{yu2025dapoopensourcellmreinforcement}. Additionally, we analyze the training stability, i.e., performance degrades or plateaus over extended runs.
These controlled variations help isolate the impact of each mechanism and offer insights into both the robustness and limitations of our current setup.

\subsection{Rollout Sampling Temperature}
\label{sec:sampling_temp}

\begin{figure}[ht] 
    \centering
    \begin{subfigure}[b]{0.3\textwidth}
        \includegraphics[width=\textwidth]{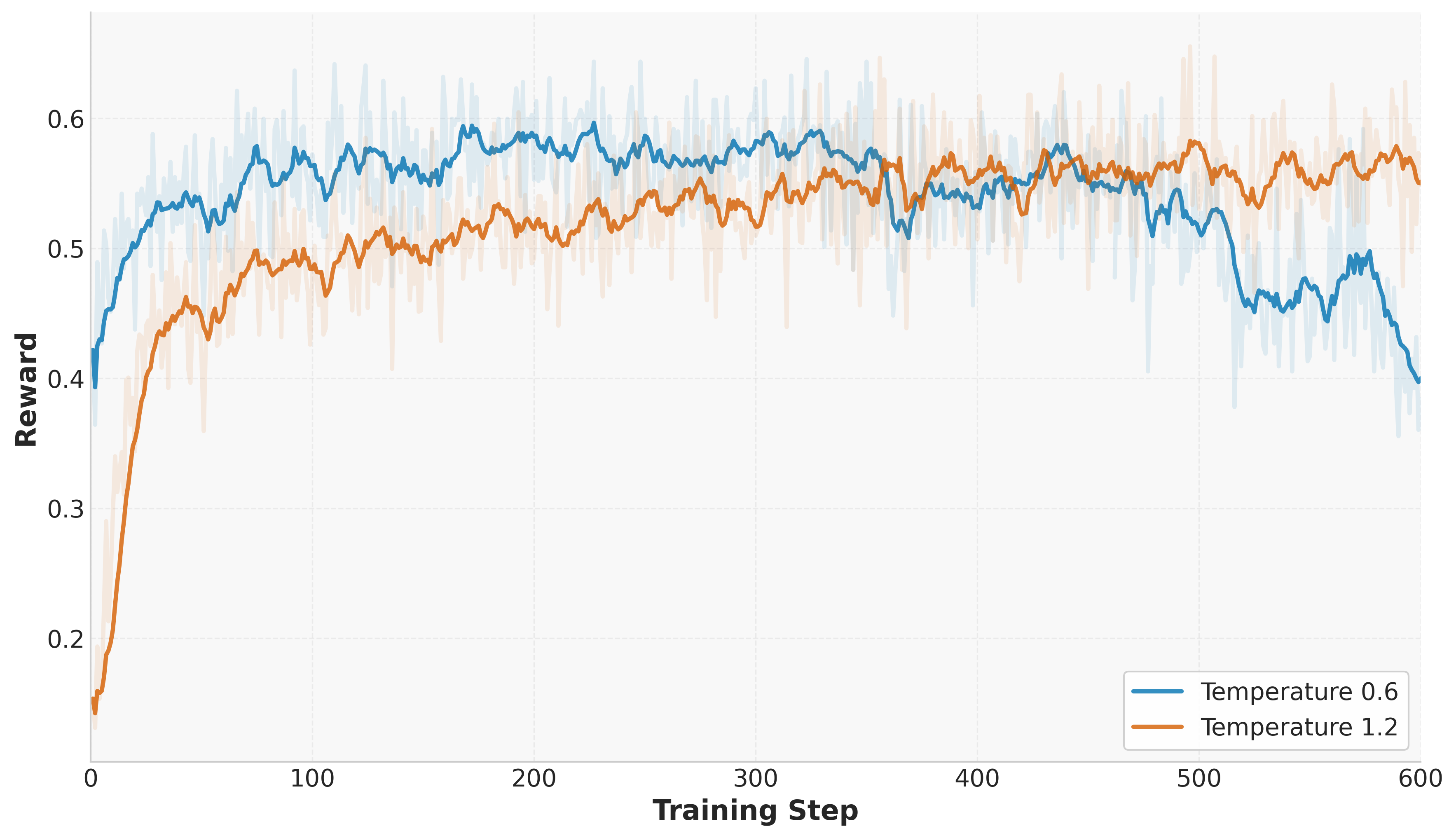}
        \caption{Reward with Run 1.}
    \end{subfigure}
    \hfill
    \begin{subfigure}[b]{0.3\textwidth}
        \includegraphics[width=\textwidth]{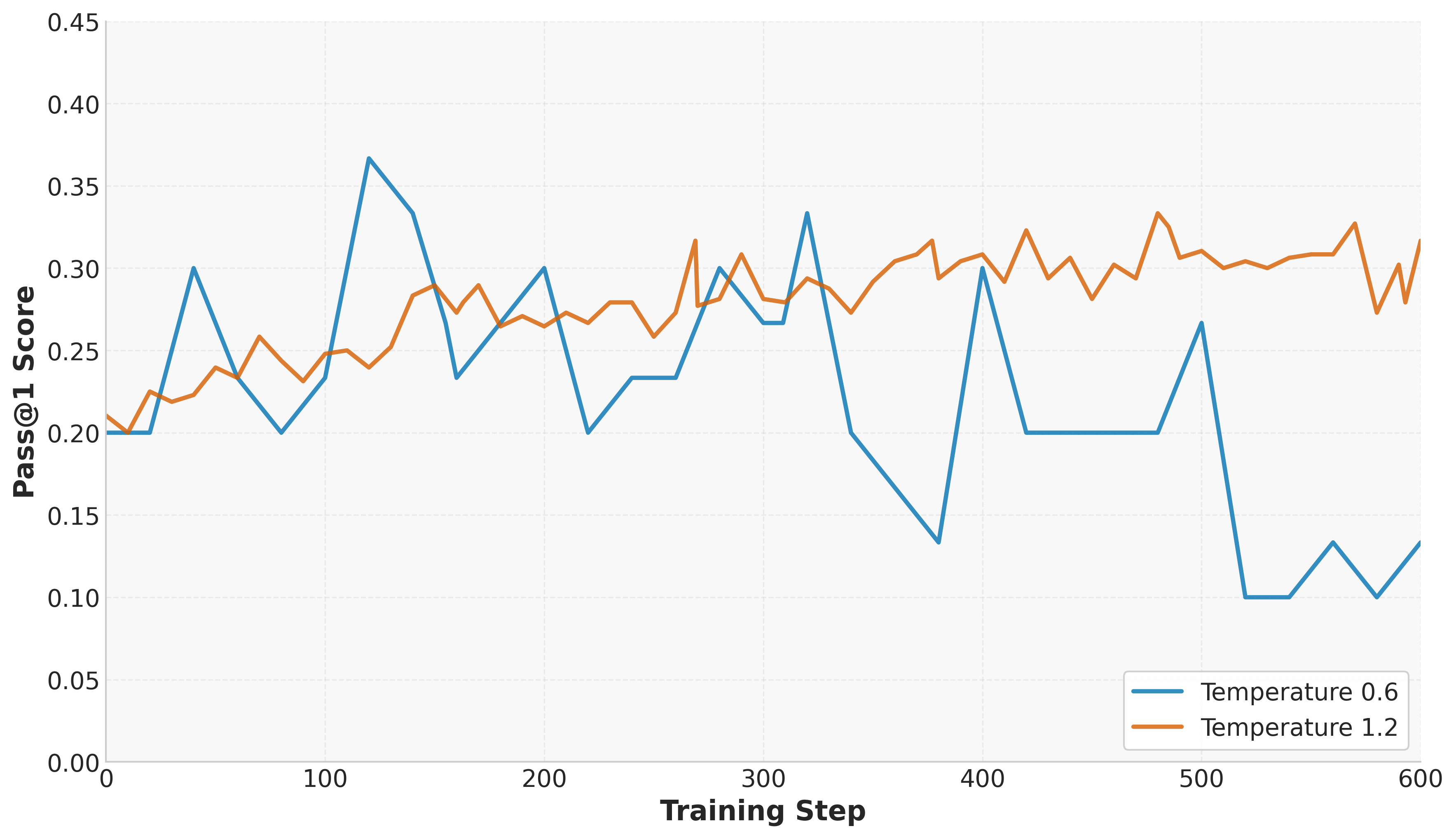}
        \caption{AIME2024 score with Run 1.}
    \end{subfigure}
    \hfill
    \begin{subfigure}[b]{0.3\textwidth}
        \includegraphics[width=\textwidth]{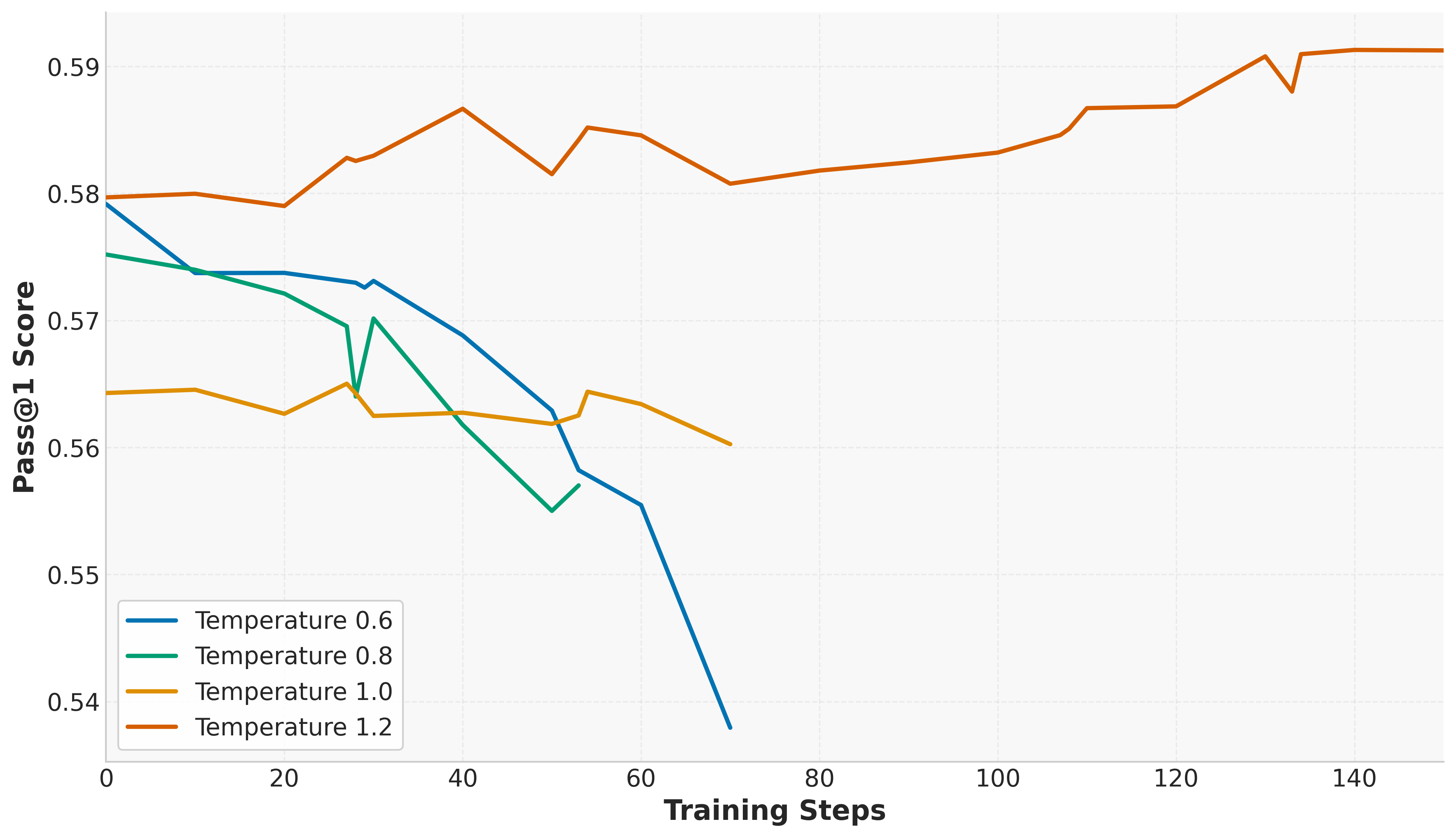}
        \caption{Validation score with Run 7.}
    \end{subfigure}

    \caption{Ablation study on rollout sampling temperature during early- (Run 1) and late-stage (Run 7) training.}
    \label{fig:ablation_temperature}
\end{figure}

We conduct ablations on the rollout sampling temperature to study its effect on training dynamics and 
keep performance metrics
shown in~\Cref{fig:ablation_temperature}. Sampling temperature plays a crucial role in shaping the distribution of rollout candidates, thereby influencing both reward signal quality and model learning behavior. Our study includes two separate evaluations: one during early-stage training (Run 1) and another during late-stage training (Run 7), before extending the context window.

\paragraph{Early-stage Training Ablation (Run 1)}
In the early reinforcement learning phase, we experiment with different sampling temperatures.
We find that training with a low temperature (e.g., 0.6) leads to early instability. Both the average reward and downstream performance on AIME2024 degrade significantly, likely due to reduced diversity in rollouts and overexploitation of suboptimal generation patterns. In contrast, using a higher temperature of 1.2 yields more stable training. Although initial rewards are lower compared to temperature 0.6, they improve steadily and eventually surpass the low-temperature setting. The accuracy score of AIME2024 follow a similar trend to the reward as the average reward in term of sampling temperature, indicating that more diverse rollouts support better generalization and learning at this stage.

\paragraph{Late-stage Training Ablation (Run 7)}
We further evaluate temperature settings during Run 7.
After extensive training and increased rollout counts, the model is already strong.
Conventional intuition might suggest that lower temperatures would be more effective in this context to leverage more exploitation over exploration. 
However, our ablation shows that high temperatures again yield better reward progression and validation stability than lower temperatures. The high-temperature rollouts prevent mode collapse and support continued progress, even when performance gains begin to plateau.

These results underscore the importance of temperature tuning throughout training. Lower temperatures may prematurely limit exploration, especially early in training, while properly higher temperatures enable broader behavioral exploration and improved reward optimization. We note that this observation may be model-dependent, and transferring it to other models may require further analysis.

\subsection{Decoupled Clip and Dynamic Sampling}
\label{sec:clip_ratio_ablation_study}
\begin{figure}[ht] 
    \centering
    \begin{subfigure}[b]{0.3\textwidth}
        \includegraphics[width=\textwidth]{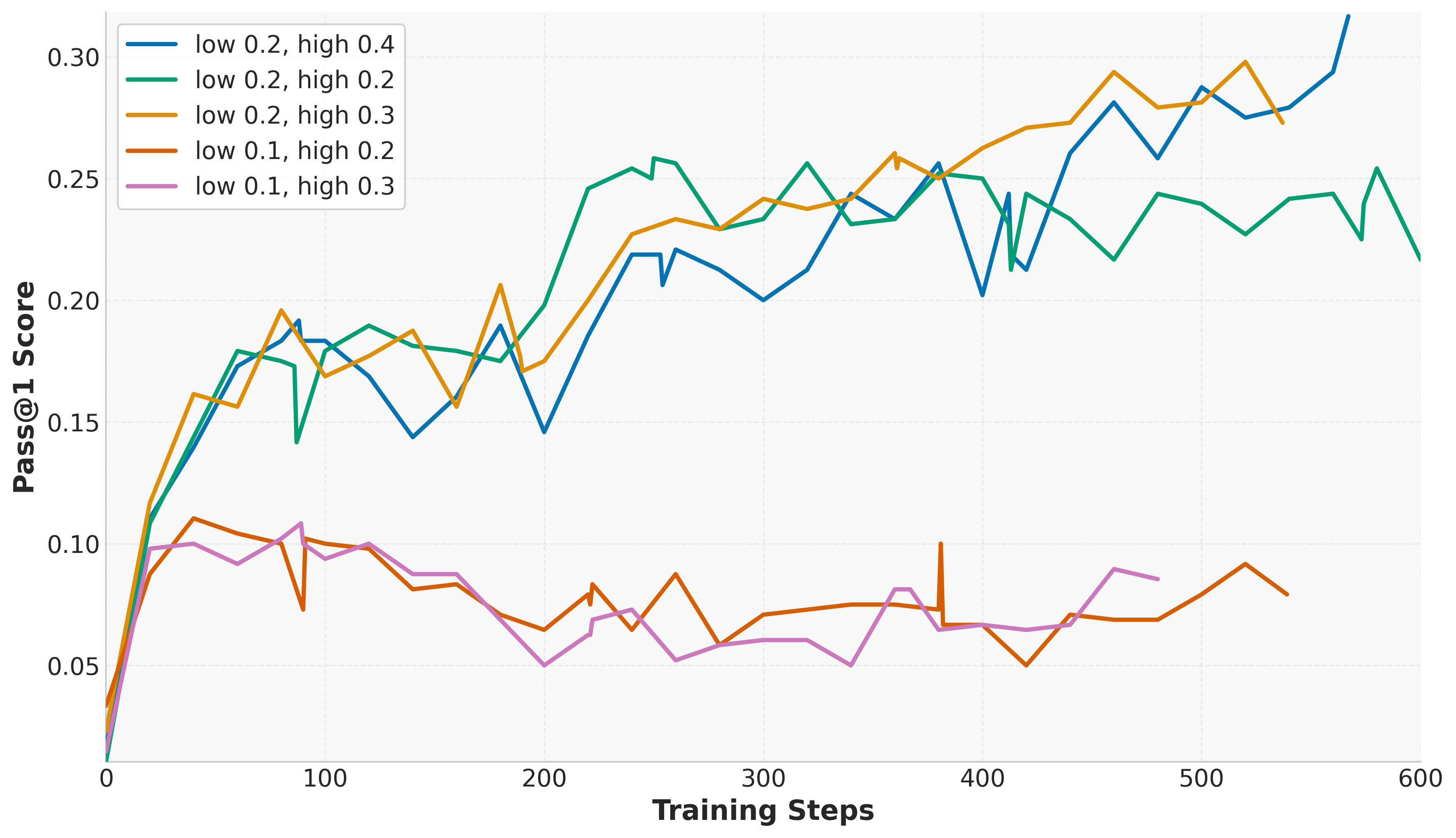}
        \caption{AIME2024 scores with different clipping coefficient.}
    \end{subfigure}
    \hfill
    \begin{subfigure}[b]{0.3\textwidth}
        \includegraphics[width=\textwidth]{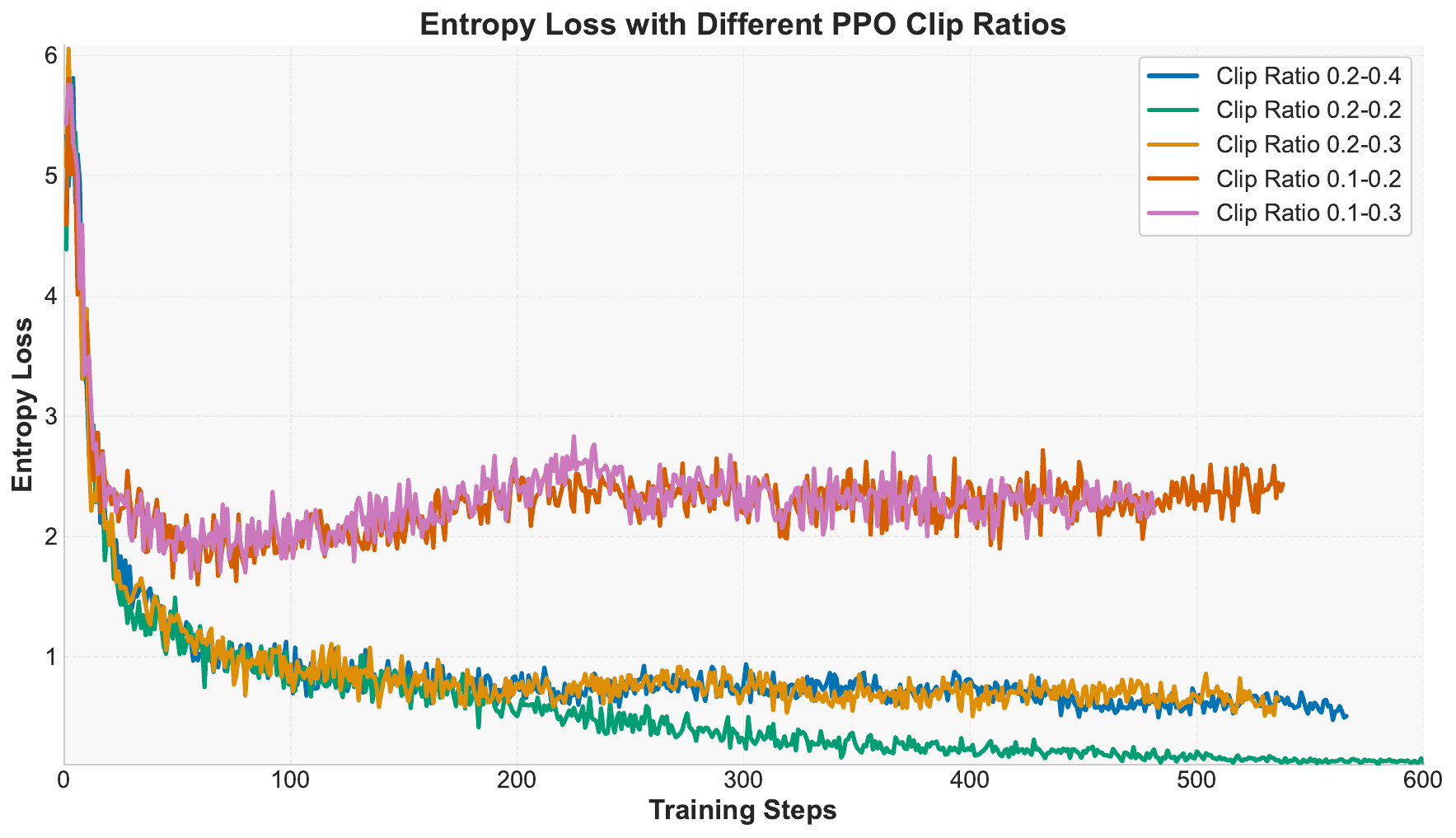}
        \caption{Entropy with different clipping coefficient.}
    \end{subfigure}
    \hfill
    \begin{subfigure}[b]{0.3\textwidth}
        \includegraphics[width=\textwidth]{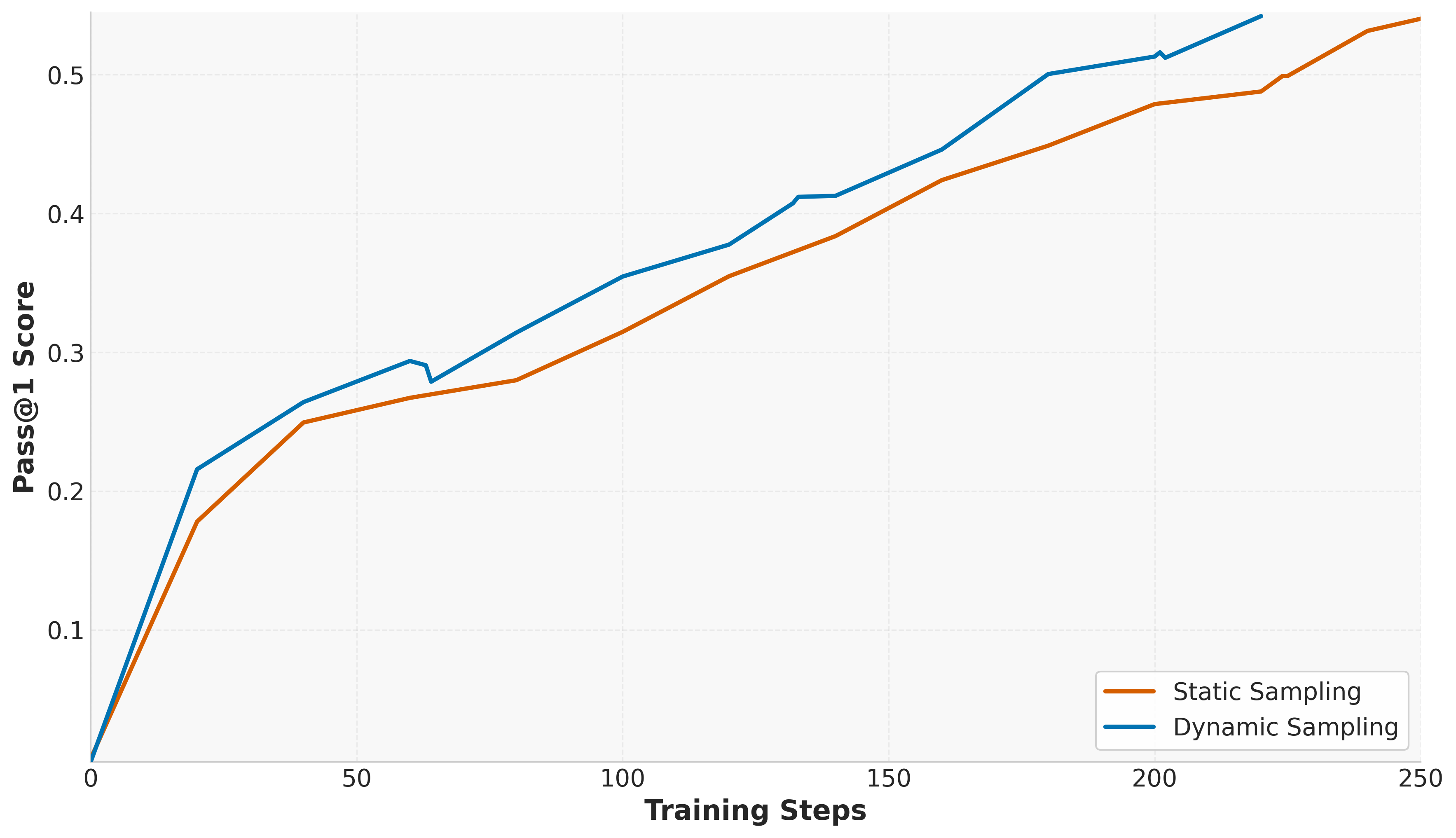}
        \caption{Validation scores with dynamic sampling.}
    \end{subfigure}

    \caption{Ablation study on decoupled clipping and dynamic sampling~\cite{yu2025dapoopensourcellmreinforcement}. Both techniques demonstrate clear improvements over standard GRPO~\cite{shao2024deepseekmath} in our experiments.}
    \label{fig:ablation_dapo}
\end{figure}


We investigate the impact of decoupled clipping by varying its lower and upper thresholds, $\epsilon_{low}$ and $\epsilon_{high}$, as illustrated in~\Cref{fig:ablation_dapo}. Our results show that setting $\epsilon_{low} = 0.2$ outperforms more conservative choices like $0.1$, indicating that more aggressively down-weighting actions with negative advantage can facilitate learning. Additionally, increasing the upper threshold to $\epsilon_{high} = 0.4$ led to further improvements, mitigating entropy collapse and yielding the highest validation performance overall.

Interestingly, we observed that applying a smaller $\epsilon_{low}$ led to higher policy entropy. This is because a smaller $\epsilon_{low}$ discourages the policy from strongly penalizing actions with negative advantage, effectively flattening the action distribution and encouraging broader exploration. However, this increased entropy did not translate into improved learning outcomes, emphasizing that excessive exploration alone is insufficient, and that a careful balance between exploration and exploitation is crucial for stable and effective training.

Finally, we verified that dynamic sampling led to faster improvements than static sampling by filtering out prompts with zero advantage. This increases 
reward signal density 
in each batch, thereby improving the overall sample efficiency of training.

\subsection{Resetting Reference Policy}

\begin{figure}[ht] 
    \centering
    \begin{subfigure}[b]{0.48\textwidth}
        \includegraphics[width=\textwidth]{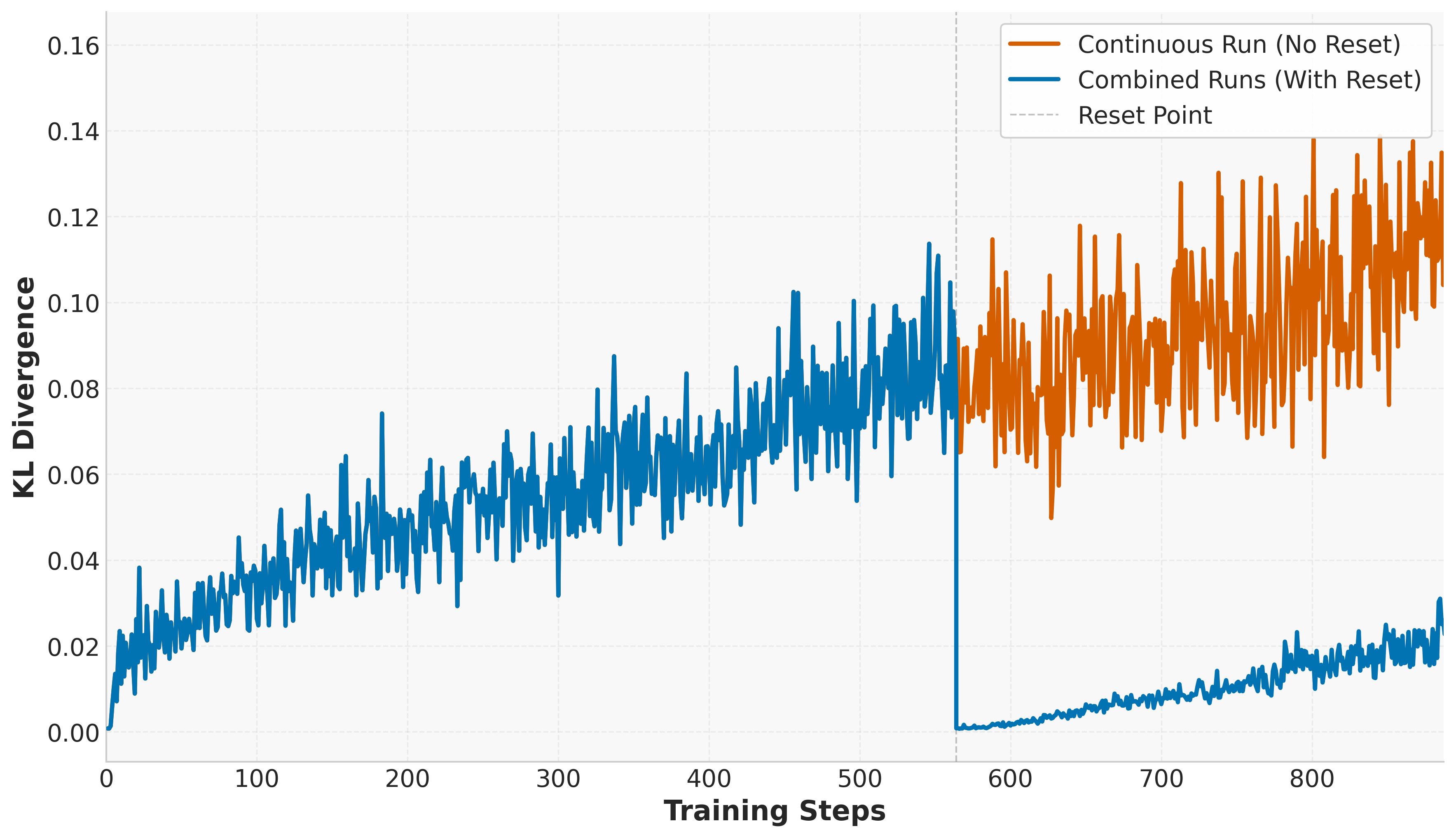}
        \caption{KL divergence.}
    \end{subfigure}
    \hfill
    \begin{subfigure}[b]{0.48\textwidth}
        \includegraphics[width=\textwidth]{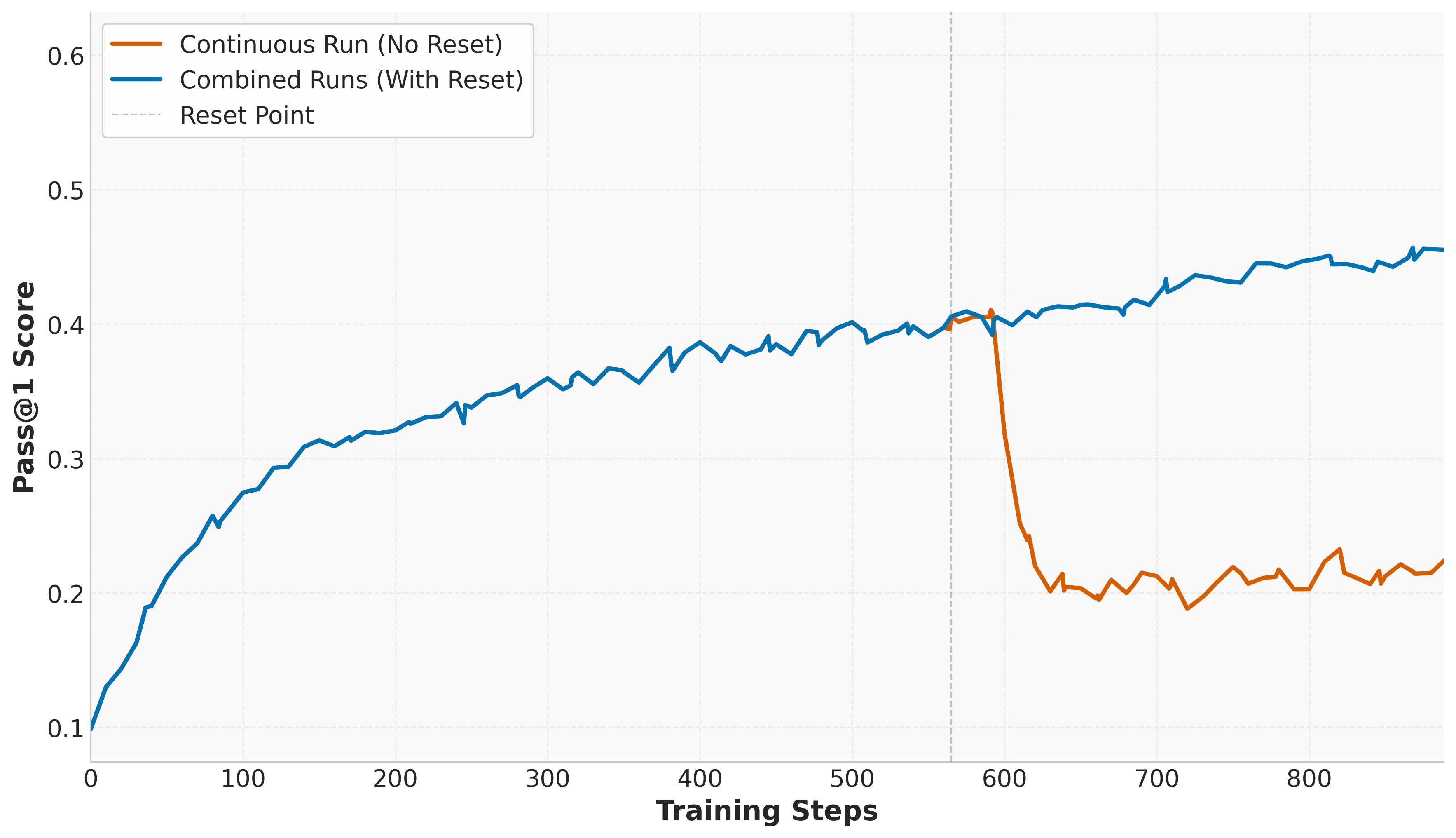}
        \caption{Codeforces scores.}
    \end{subfigure}

    \caption{Extended training sometimes becomes unstable. 
    In our case, extending Run 1 causes a sudden degradation on Codeforces validation scores. 
    In other cases, improvement plateaus with a sudden increase in KL divergence. 
    We perform a hard reset of training, resetting the reference policy used for KL divergence calculation as well as optimizer states.}
    \label{fig:reset}
\end{figure}

We observe that extending training with the same setting and training hyperparameters does not always lead to consistent improvements. As shown in~\Cref{fig:reset}, if we simply extend the training of Run 1, it encounters a sharp drop in validation scores on the Codeforces benchmark. In other scenarios, performance gains plateau and the KL divergence against the reference policy spikes unexpectedly. 
To mitigate these issues, we implement a hard reset strategy, refreshing both the reference policy (used for KL divergence computation) and the optimizer states, to restore stability and enable further effective training.

\subsection{Mitigating Entropy Collapse}

\begin{figure}[ht] 
    \centering
    \begin{subfigure}[b]{0.48\textwidth}
        \includegraphics[width=\textwidth]{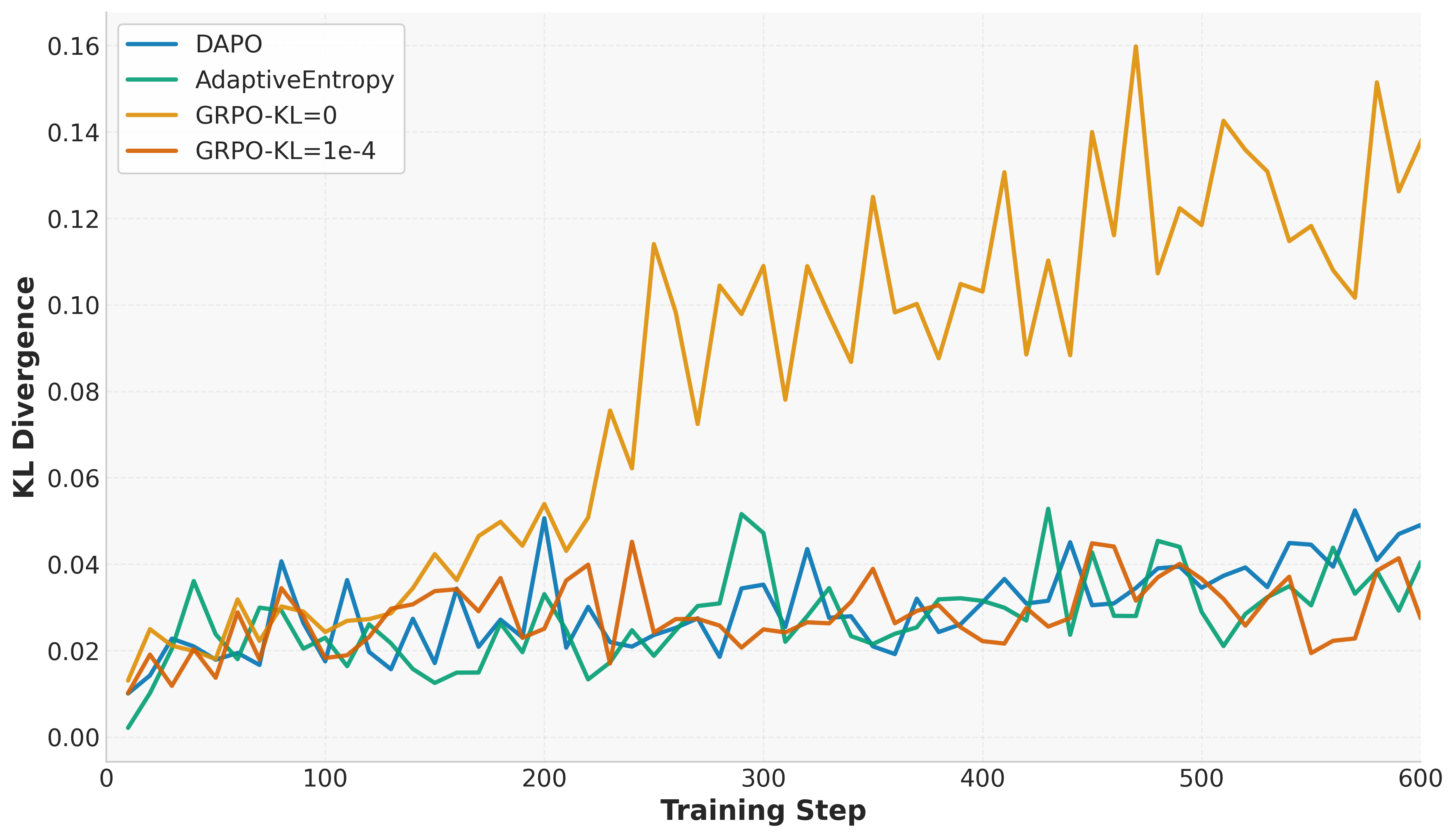}
        \caption{KL divergence.}
    \end{subfigure}
    \hfill
    \begin{subfigure}[b]{0.48\textwidth}
        \includegraphics[width=\textwidth]{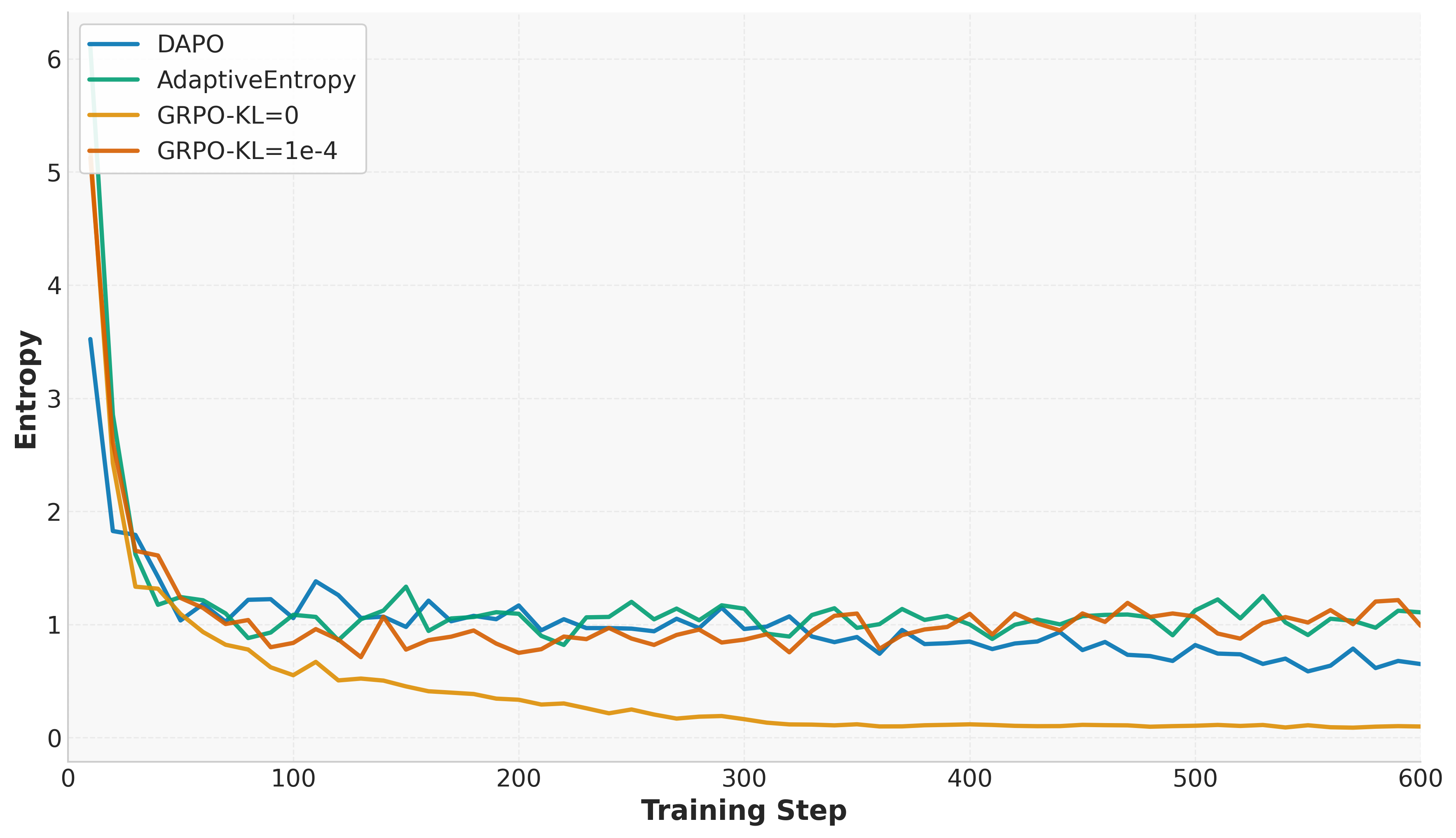}
        \caption{Entropy.}
    \end{subfigure}

    \vspace{0.5cm}

    \begin{subfigure}[b]{0.48\textwidth}
        \includegraphics[width=\textwidth]{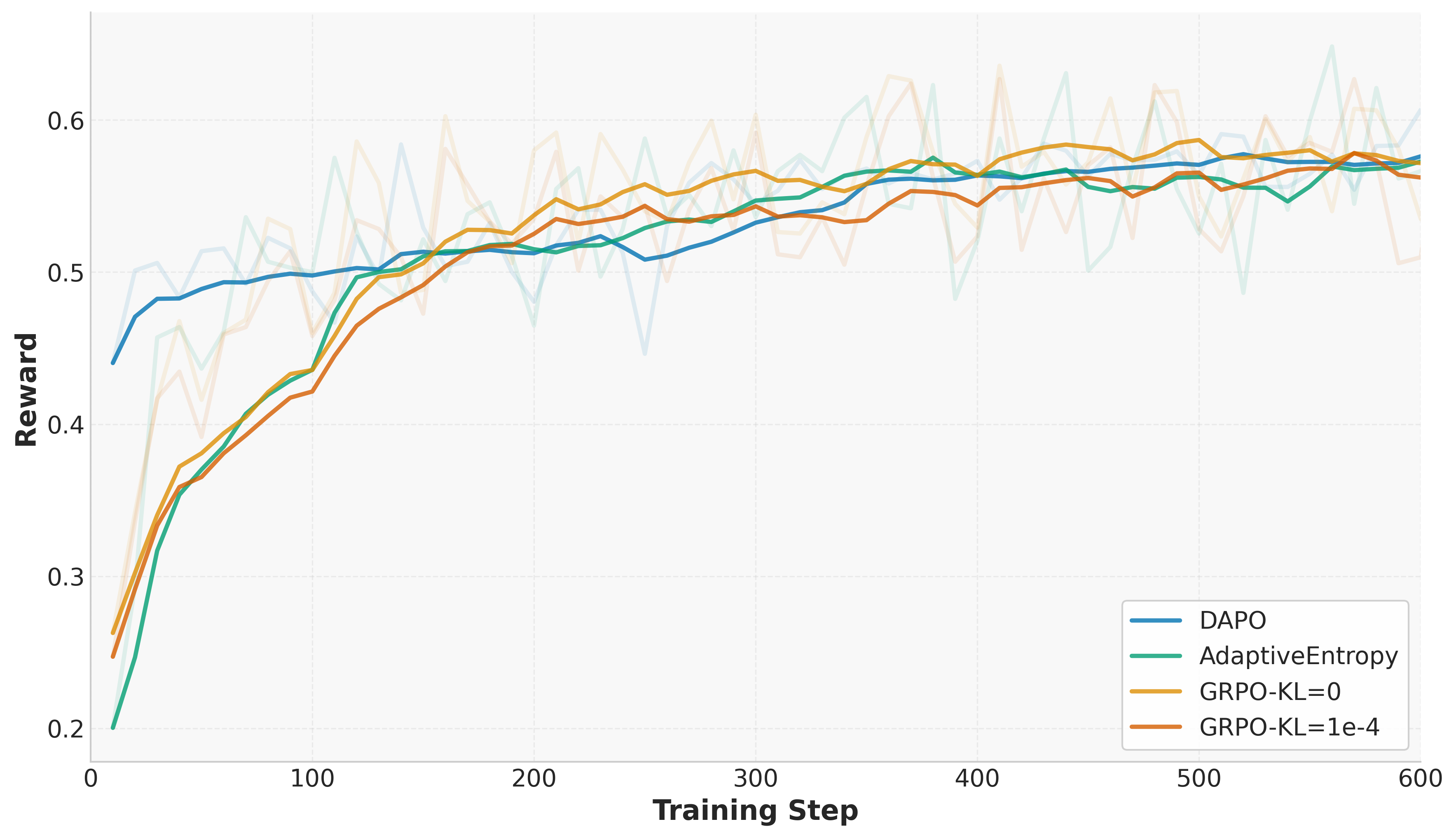}
        \caption{Reward.}
    \end{subfigure}
    \hfill
    \begin{subfigure}[b]{0.48\textwidth}
        \includegraphics[width=\textwidth]{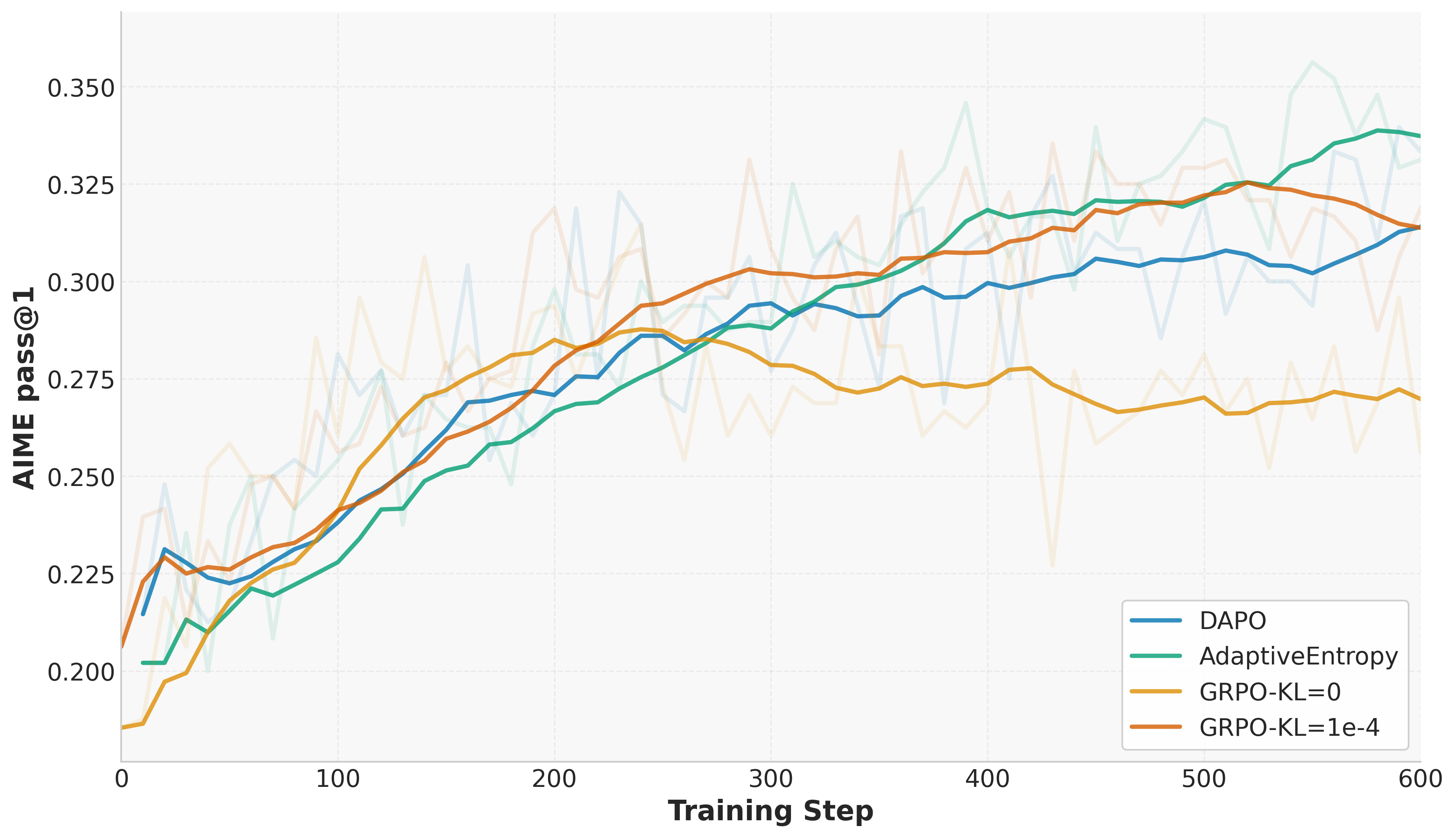}
        \caption{AIME2024 scores.}
    \end{subfigure}

    \caption{Training dynamics under different entropy mitigation strategies.}
    \label{fig:ablation_entropy_collapse}
\end{figure}

Reinforcement learning often suffers from entropy collapse, where the model's output distribution becomes too concentrated early in training, leading to a rapid drop in entropy. This can hinder exploration and bias advantage estimation, ultimately stalling learning. 
Figure~\ref{fig:ablation_entropy_collapse} illustrates the phenomenon: without intervention, the model's output entropy drops sharply early in training and remains low, signaling a premature convergence to narrow output distributions. 
To better understand and mitigate entropy collapse during reinforcement learning, we conducted a series of ablation studies evaluating the training dynamics under different strategies.
We investigated the following settings:
\begin{itemize}

\item \textbf{GRPO-KL=0.} We employ standard GRPO training without KL divergence penalty or entropy bonus.
\item \textbf{GRPO-KL=1e-4} We apply a small KL regularization term to control the divergence between the online policy and a reference policy. This helps prevent the model from drifting too far and encourages continued exploration.
\item \textbf{DAPO.} We adopt the `clip-higher' technique from DAPO~\citep{yu2025dapoopensourcellmreinforcement}, which upweights the probabilities of lower-ranked tokens during sampling. 

\item \textbf{AdaptiveEntropy.} Inspired by parallel work~\citep{skywork-or1-2025}, we also experiment with an adaptive entropy strategy, where the entropy bonus coefficient is dynamically enabled or disabled based on training statistics. Since tuning a fixed entropy coefficient is notoriously brittle, this approach sets a target entropy and selectively applies the bonus to maintain a desired level of output diversity.
\end{itemize}

We observe that all three methods are effective in mitigating entropy collapse to varying degrees. The adaptive entropy strategy introduces the most flexibility, but requires careful calibration of the target entropy and 
adds one more sensitive hyperparameter to tune.
In contrast, DAPO and KL penalty offer a more conservative and robust solution, especially when training from a capable base model. Based on these findings, we adopt a combination of DAPO and a KL penalty in our final training setup. The KL term not only contributes to entropy preservation but also improves training stability by preventing runaway divergence from the reference model. Together, these methods strike a balance between preserving exploration and maintaining coherent, high-quality generations throughout prolonged training.

\section{Conclusion}
In this technical report, we presented a comprehensive investigation of prolonged reinforcement learning for reasoning-focused language models, identifying critical components that enable stable and effective training across diverse tasks. Our work demonstrates that through careful algorithm design, including decoupled clipping, dynamic sampling, controlled KL regularization, and periodic reference policy resets, even small-scale models can achieve substantial reasoning improvements without the computational demands of larger architectures. These techniques yielded significant performance gains over the baseline DeepSeek-R1-Distill-Qwen-1.5B model across multiple domains, including mathematics (+14.7\%), coding (+13.9\%), logic puzzles (+54.8\%), STEM reasoning (+25.1\%), and instruction-following (+18.1\%). Importantly, our approach maintained competitive performance with domain-specialized models despite being trained on a broader range of tasks, suggesting that properly implemented reinforcement learning and prolonged training can effectively bridge the gap between general-purpose models and specialized reasoning systems. By open-sourcing our model and sharing our training methodology, we hope to facilitate further advancements in alignment, optimization, and reasoning within resource-efficient language models.

\appendix
\section{Contributors}
Mingjie Liu, Shizhe Diao, Jian Hu, Ximing Lu, Xin Dong, Hao Zhang, Alexander Bukharin, Shaokun Zhang, Jiaqi Zeng, Makesh Narsimhan Sreedhar, Gerald Shen, David Mosallanezhad, Di Zhang, Jonas Yang, June Yang, Oleksii Kuchaiev, Guilin Liu, Zhiding Yu, Pavlo Molchanov, Yejin Choi, Jan Kautz, Yi Dong

\newpage
{
  \small
  \bibliographystyle{unsrt}
  \bibliography{main_tech_report}
}


\end{document}

%% file: math_commands.tex
\def\eqref#1{equation~\ref{#1}}

\def\1{\bm{1}}

\DeclareMathAlphabet{\mathsfit}{\encodingdefault}{\sfdefault}{m}{sl}
\SetMathAlphabet{\mathsfit}{bold}{\encodingdefault}{\sfdefault}{bx}{n}



%% file: main_method.tex
\section{Approach}
\label{sec:method}

We begin with a brief overview of the algorithm we used to optimize our language model, GRPO~\cite{shao2024deepseekmath}. 
 Our approach integrates techniques from DAPO~\cite{yu2025dapoopensourcellmreinforcement}, such as decoupled clipping and dynamic prompt sampling. Finally, to further address entropy collapse and training instatility, we apply a KL divergence penalty and introduce a periodic reset of the reference policy. These components together enable stable training over extended durations and repeated epochs on the same dataset without degrading performance.

\subsection{Background: Group Relative Policy Optimization}
We adopt Group Relative Policy Optimization (GRPO)~\cite{shao2024deepseekmath} as the core RL algorithm. Compared with Proximal Policy Optimization (PPO)~\cite{schulman2017proximalpolicyoptimizationalgorithms}, it removes the value model and instead use baseline estimates based on group scores. Formally the GRPO maximizes the following objective:
\begin{equation}
  \mathcal{L}_{\text{GRPO}}(\theta) = \mathbb{E}_{\tau \sim \pi_\theta} \Bigg[
\min\Big(
r_\theta(\tau) A(\tau), \\
\quad \text{clip}(r_\theta(\tau), 1 - \epsilon, 1 + \epsilon) A(\tau)
\Big)
\Bigg],  
\label{equ:grpo}
\end{equation}
where $\tau$ is the response sampled from the current policy $\pi_\theta$. 
$r_\theta(\tau)=\frac{\pi_\theta(\tau)}{\pi_{old}(\tau)}$ is the probability ratio between the current policy and old policy before each actor update. The advantage used in GRPO foregoese the critic model of PPO, and instead estimates baseline from group scores $\{R_i\}_{i\in G(\tau)}$:
\begin{equation}
    A(\tau) = \frac{R_\tau - mean(\{R_i\}_{i\in G(\tau)})}{std(\{R_i\}_{i\in G(\tau)})}.
\end{equation}

\subsection{Prolonged Reinforcement Learning}
\subsubsection{Mitigating Entropy Collapse}

A key challenge in prolonged policy optimization is entropy collapse, a phenomenon where the model’s output distribution becomes overly peaked early in training, resulting in sharply reduced entropy. When entropy collapses, the policy prematurely commits to a narrow set of outputs, severely limiting exploration. This is particularly detrimental in methods like GRPO, where the learning signal depends on having a diverse set of sampled outputs to effectively estimate relative advantages. Without sufficient exploration, policy updates become biased, leading to stagnation in training.

A common mitigation strategy is to increase the sampling temperature during rollouts. However, we find that this approach merely delays the onset of entropy collapse rather than preventing it entirely, as entropy continues to decline steadily throughout training. As detailed in our temperature ablation study in Section~\ref{sec:sampling_temp}, we used a high rollout temperature to promote exploration by boosting the initial entropy.

\subsection{Decoupled Clip and Dynamic Sampling Policy Optimization (DAPO)}

To address entropy collapse, we adopt several components from the DAPO algorithm~\cite{yu2025dapoopensourcellmreinforcement}, which are specifically designed to maintain exploration and output diversity. First, DAPO introduces decoupled clipping, where the lower and upper clipping bounds in the PPO objective are treated as separate hyperparameters:
\begin{equation}
\text{clip}(r_\theta(\tau), 1 - \epsilon_{low}, 1 + \epsilon_{high}).
\end{equation}
By setting a higher value for $\epsilon_{high}$, the algorithm promotes `clip-higher', uplifting the probabilities of previously unlikely tokens and encouraging broader exploration. We find that this modification helps retain entropy and reduces premature mode collapse.

Additionally, DAPO employs dynamic sampling, filtering out prompts for which the model consistently succeeds or fails (i.e., accuracy 1 or 0), as these provide no learning signal. This focus on intermediate difficulty examples further helps maintain a diverse and stable learning signal during training.

\subsubsection{KL Regularization and Reference Policy Reset}
\label{sec:kl_reset}

While DAPO and temperature adjustment help slow entropy collapse, we find that explicit regularization via a KL divergence penalty provides a stronger and more stable solution. Specifically, we incorporate a KL penalty between the current policy $\pi_\theta$ and a reference policy $\pi_{ref}$:
\begin{equation}
L_{KL-RL}(\theta) = L_{GRPO}(\theta) - \beta D_{KL}(\pi_\theta || \pi_{ref}),
\end{equation}
where the following unbiased estimator is commonly used\cite{noauthor_approximating_nodate}:
\begin{equation}
    D_{KL}(\pi_\theta||\pi_{ref}) = \frac{\pi_{ref}(\tau)}{\pi_\theta(\tau)} - log\frac{\pi_{ref}(\tau)}{\pi_\theta(\tau)} - 1.
\end{equation}
This penalty not only helps maintain entropy but also serves as a regularizer to prevent the online policy from drifting too far from a stable reference, stabilizing learning and mitigating overfitting to spurious reward signals.

Recent works~\cite{yu2025dapoopensourcellmreinforcement, deepcoder2025, yue2025vapoefficientreliablereinforcement, skywork-or1-2025} have argued for the removal of the KL penalty, citing that models naturally diverge during training on chain-of-thought reasoning tasks. We observe that this perspective often applies to base models prior to any supervised fine-tuning. In contrast, we begin from a well-initialized checkpoint (DeepSeek-R1-Distill-Qwen-1.5B) already capable of generating coherent CoT outputs. In this context, retaining a KL penalty is still beneficial for both stability and sustained entropy.

We further observe that as training progresses, the KL term may increasingly dominate the loss, leading to diminishing policy updates. To alleviate this, we introduce a simple yet effective technique: \textit{reference policy reset}. Periodically, we hard-reset the reference policy $\pi_{ref}$ to a more recent snapshot of the online policy $\pi_\theta$, and reinitialize the optimizer states. This allows the model to continue improving while maintaining the benefits of KL regularization. We apply this reset strategy throughout training to avoid premature convergence and encourage prolonged training.